\documentclass[journal]{IEEEtran}
\IEEEoverridecommandlockouts
\usepackage[ruled,vlined,linesnumbered]{algorithm2e}
 \usepackage{xcolor}
 \usepackage{subfigure}
 \usepackage{graphicx}
\usepackage{array, makecell}
\usepackage{capt-of,etoolbox}
\usepackage{amsfonts}
\usepackage{listings}
\usepackage{xspace}
\usepackage{multirow}
\usepackage{array}
\usepackage{booktabs} 
\usepackage{mathtools}
\usepackage[ruled,vlined]{algorithm2e}
\usepackage{cite}
\usepackage{algorithmic} 
\usepackage{titlesec}
\usepackage{amsmath}
\usepackage[super]{nth}
\usepackage[export]{adjustbox}
\usepackage[bottom]{footmisc}
\usepackage{hyperref}

\makeatletter
\def\mathcolor#1#{\@mathcolor{#1}}
\def\@mathcolor#1#2#3{%
  \protect\leavevmode
  \begingroup
    \color#1{#2}#3%
  \endgroup
}
\titlespacing{\subsubsection}{0pt}{\parskip}{-\parskip}
\makeatother
\newcolumntype{P}[1]{>{\centering\arraybackslash}p{#1}}
\def\BibTeX{{\rm B\kern-.05em{\sc i\kern-.025em b}\kern-.08em
 T\kern-.1667em\lower.7ex\hbox{E}\kern-.125emX}}
 \def\BibTeX{{\rm B\kern-.05em{\sc i\kern-.025em b}\kern-.08em
    T\kern-.1667em\lower.7ex\hbox{E}\kern-.125emX}}

\DeclarePairedDelimiter{\ceil}{\lceil}{\rceil}
\begin{document}

\title{\huge {Real Time Collision Avoidance with \\GPU-Computed Distance Maps}}

 \author{
 \IEEEauthorblockN{Wendwosen Bellete Bedada, Gianluca Palli}\\
  \thanks{Wendwosen Bellete Bedada and Gianluca Palli are with the DEI - Department of Electrical, Electronic and Information Engineering, University of Bologna, Viale Risorgimento 2, 40136 Bologna, Italy.} 
 \IEEEauthorblockN{DEI - Department of Electrical, Electronic and
  Information Engineering} \IEEEauthorblockN{University of Bologna,
  Viale Risorgimento 2, 40136 Bologna, Italy} 
  \thanks{This work was supported by the European Commission’s Horizon
  2020 Framework Programme with the project REMODEL - Robotic
  technologies for the manipulation of complex deformable linear
  objects - under grant agreement No 870133.}
  \thanks{Corresponding
  author: \textit{gianluca.palli@unibo.it}}
}

\maketitle

\begin{abstract}
This paper presents reactive obstacle and self-collision avoidance of redundant robotic manipulators within real time kinematic feedback control using GPU-computed distance transform. The proposed framework utilizes discretized representation of the robot and the environment to calculate 3D Euclidean distance transform for task-priority based kinematic control. The environment scene is represented using a 3D GPU-voxel map created and updated from a live pointcloud data while the robotic link model is converted into a voxels offline and inserted into the voxel map according to the joint state of the robot to form the self-obstacle map. The proposed approach is evaluated using the Tiago robot, showing that all obstacle and self collision avoidance constraints are respected within one framework even with fast moving obstacles while the robot performs end-effector pose tracking in real time. A comparison of related works that depend on GPU and CPU computed distance fields is also presented to highlight the time performance as well as accuracy of the GPU distance field. 
\end{abstract}

\begin{IEEEkeywords}
Collision Avoidance, self-collision and  Obstacle Avoidance, Task Priority, GPU Distance Field 
\end{IEEEkeywords}

\section{Introduction}

Real time collision avoidance are shown to be relevant for robots in unknown dynamic environments, for example in unmanned aerial vehicles and mobile manipulation applications. 
They are based on reactive motion control which allow them to be employed within real time feedback loop and are suitable for applications with unstructured workspaces. As the paradigm of human robot interaction (HRI) shifts from complete autonomy to collaboration, Real-time reactive collision avoidance should accommodate the presence of humans in the vicinity of robots without posing danger to their safety. If human operators are considered in the robot environment, the safety requirements are even more stringent, since human movements can be fast and difficult to predict. In this regard, the speed at which the environment scene is updated, minimum obstacle distance and control commands are computed  play crucial factor in utilizing real time collision avoidance in the context of HRI.  

The perception pipelines dedicated to capturing the scene from live sensors and feed the robot control algorithms with the important information about obstacles usually consume significant (up to 90\%) computational power~\cite{pan2010g}. To minimize the latency during collision checking, the environment model is usually restricted to 2.5D and the robot model undergoes significant simplification in their representation \cite{C144, di2018safety}. 
However, the advance in computing capability, particularly the emergence of high performance GPU's has accelerated the computation time thereby allowing a complete 3D representation of the environment and robots. In this aspect, GPU-Voxels is an open source library for CUDA based
Graphics Processing Units (GPU) which allows massively parallel computation of collision checking in 3D environments \cite{A1} and offers constant runtime regardless of the occupancy density in the environment. It is also computationally inefficient and unnecessary to compute minimum obstacle distance to every point on the robot body, therefore requiring a simplified but correct robot model approximations. 
\begin{figure}[t]
 \centering
 \subfigure[Tiago robot voxel model.\label{fig:rob_env}]{ \includegraphics[width=0.45\columnwidth]{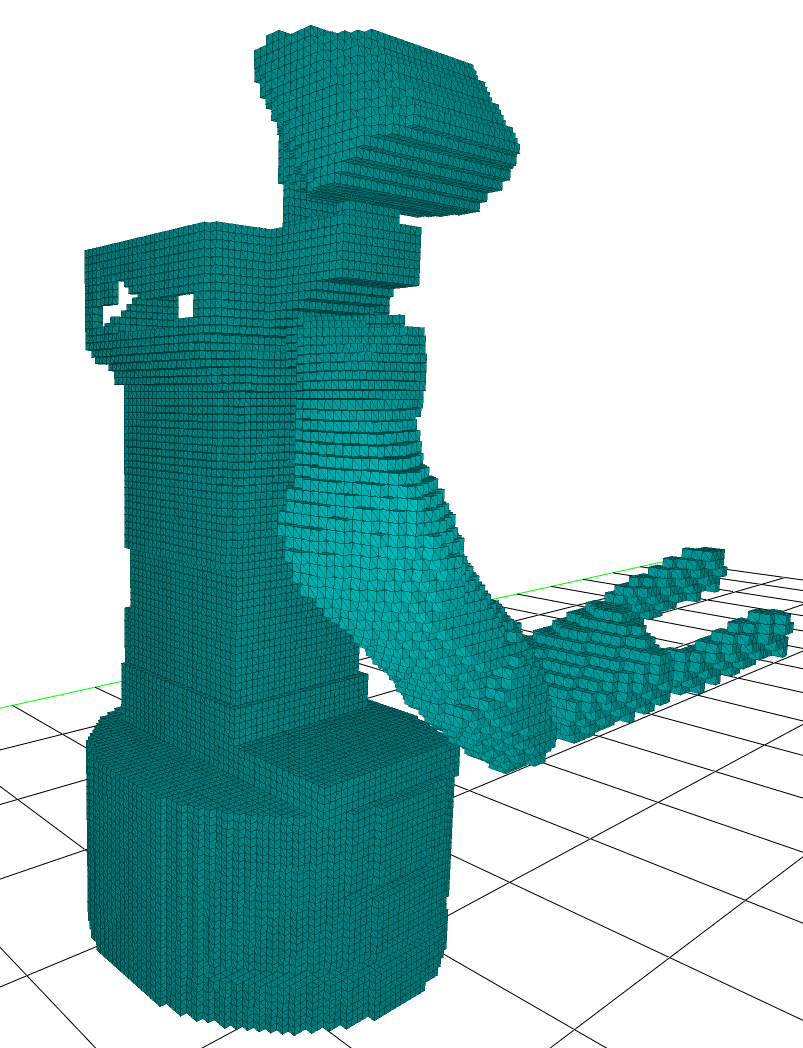}
 }
 \subfigure[Obstacle links and bounding sphere approximation of the manipulator.\label{fig:centers}]{
 \includegraphics[width=0.45\columnwidth]{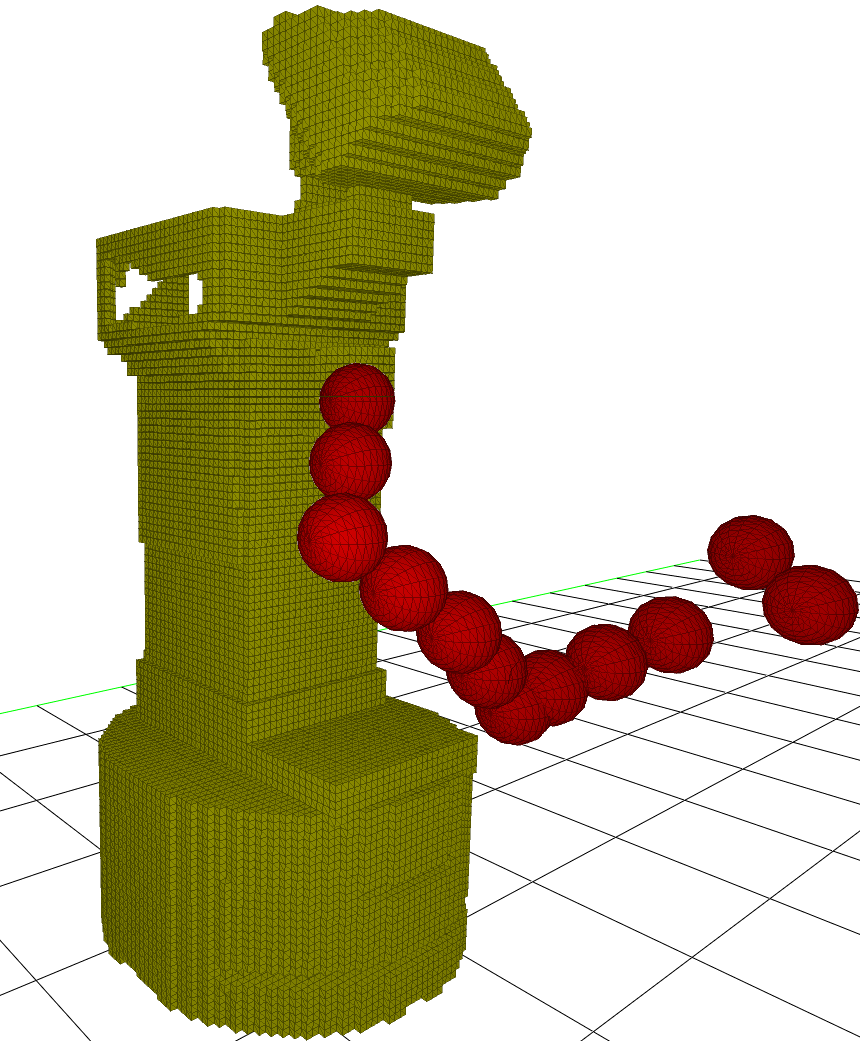}
 }
 \caption{Representation of the Tiago robot in the scene: (a) links are offline voxelized and inserted according to the joint state of the robot; (b) the arm of the robot is approximated with spherical volumes, the body highlighted in yellow is used as an obstacle for self-collision avoidance.}
\end{figure}
In addition to the requirement for computational efficiency, real time collision avoidance relies on an algorithm that generate control directions that avoids obstacles. Task priority control is commonly used in this scenario since it allows to activate and deactivate tasks with higher priority, e.g. collision avoidance, only when they are really needed. 

\begin{figure}
 \centering
 \subfigure[OBBs defined over robot links.\label{fig:obb}]{
  \includegraphics[width=5cm, height=3.6cm,
  keepaspectratio]{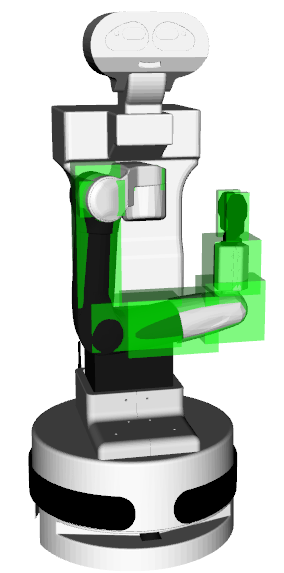}
 }
 \hspace{3mm}
  \subfigure[Spheres defined over the OBBs.\label{fig:obb_spheres}]{
  \includegraphics[width=7cm, height=3.6cm,
  keepaspectratio]{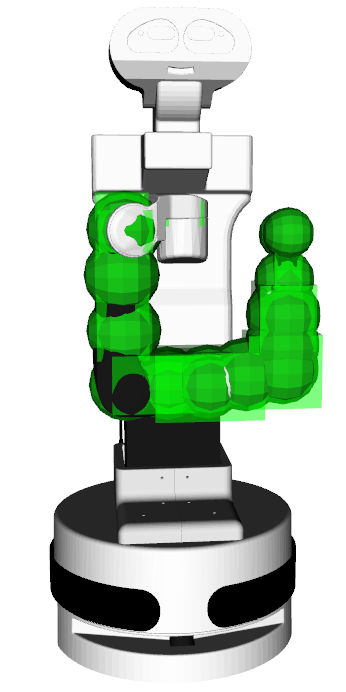}
 }\\
 \caption{ (a) OBB defined over each link; (b) OBB-Aligned minimal spherical bounding volume Bounding: computed using OBB of each link.\label{fig:real-Tiago-wm}}
 \vspace*{-3mm}
\end{figure}

In this paper, a generalized obstacle avoidance technique exploiting GPU voxels to compute 3D Euclidean Distance Transform (EDT) for obstacle and self-collision avoidance is proposed. The EDT contains the nearest obstacle information of all environment voxels which can be enquired to give the closest obstacles of any number points on the robot links. In this work, the parallelism offered by the GPU is exploited by removing the typical hierarchical representation of e.g. Octomap and directly performing distance computation on a body tight spheres as shown in Fig~\ref{fig:obb_spheres}. Therefore, instead of relying on hierarchical representations, the novelty in the proposed algorithm is that it only relies on simplified robot shape representation by means of spheres that tightly approximate the robot links shape together with high resolution distance field of the environment in the range of 1 cm, thereby allowing collision free movement even in a close vicinity of the obstacle. Task priority control is then exploited to implement the collision avoidance task. The proposed approach is evaluated on a mobile manipulator, the Tiago robot from PAL Robotics, showing that all obstacles and self collision are avoided within one single framework in real time and in presence of dynamic obstacles, while the robot simultaneously performs end-effector pose tracking. A comparison with related algorithms that depend on CPU computed distance fields is also presented to highlight the time performance as well as accuracy of the GPU distance field.  
The key contributions of this paper therefore are:
\begin{enumerate}
    \item Integration of Reactive collision avoidance with  fast  GPU-based  distance  field;
    \item A simplified robot model (OBB-Aligned minimal spherical bounding volumes) capable of exploiting the GPU computation is proposed;
    \item A novel approach for self collision avoidance that processes a voxelized model of each link on a GPU to compute minimum self collision distance;
    \item A flexible task priority controller suitable for dynamic task transition and able to handle multiple tasks including  obstacle and self-collision avoidance in one framework; 
\end{enumerate}

And finally, Practical  demonstration  of  real-time  reactive  collision avoidance with live sensor data in a dynamic 3D scene contributing to practical usage of the algorithm is demonstrated.
The remainder of the paper is organized as follows. In Section~\ref{sec:literature}, a literature review about collision avoidance and distance field computation is presented. The general framework of the proposed approach is reported in the context task-priority control using GPU-computed exact EDT together with an algorithm that unifies them in section~\ref{sec:Method}. The summary of the mathematical background for task priority control and task oriented regularization is also presented in this section. Section~\ref{sec:experiments} discusses experimental results both for obstacle and self-collision avoidance, together with a comparison with similar approaches reported in literature. Finally Section~\ref{conclusion} presents conclusion and proposes a future extension.

\section{Literature Review} \label{sec:literature}
Robot collision avoidance  task has been approached as an offline or online problem. In the offline case, the goal is to solve a motion planning problem that is free of collision~\cite{karaman2011sampling}, \cite{karaman2011anytime}, \cite{moll2015benchmarking}, \cite{koenig2002d}  whereas online collision avoidance is characterized by a reactive motion and it is susceptible to local minima. When the robot operates in a dynamic environment it has to react very fast to obstacles putting a real time requirement.  
A reactive motion planning approach is  proposed for collision avoidance in \cite{A1} using exact 3D EDT that enable online motion re-planning. In \cite{C1} a Task-Priority controller which supports set-based tasks is implemented for obstacle avoidance of an underwater vehicle using spherical collision objects generated from multi resolution Octomap for the obstacle representation. However, the current Octomap implementation only relies on Central Processing Unit (CPU) and it is therefore expected that it yields  poor real time performance \cite{C2}. 
On the other hand, GPU-implemented Octree is shown to improve  performance by lending a powerful tool to parallelly search for colliding voxels( grid points that are occupied both  by the robot and the obstacle)\cite{hermann2014unified}. A real time self collision avoidance based on Euclidean distance calculation between bounding spheres rigidly  attached to robot links using robot kinematics is reported in \cite{C12,C16}. The work in \cite{C14} by Jia Pan et al presents a real time dynamic AABB tree for fast culling as well as  direct usage of an octree for obstacle data structure representation. However, octrees are known to slow down the update from incoming pointcloud, therefore limiting the performance. A different approach based only on depth image as opposed to point clouds to compute minimum obstacle distance is reported in \cite{C143}. A recent work based on  Potential field method for real-time collision avoidance that exploits minimum distance computation between geometric primitives is reported in \cite{safeea2019line}. In this work four wearable inertial sensors are utilized per person to determine the pose of the geometric shapes approximating the shape of the person~\cite{safeea2019minimum}, thereby providing human-like reflexes. It should be noted that the approach proposed in \cite{safeea2019line} is more suited to an industrial scenario where one person with a wearable device interact with the robot. Another recent work in~\cite{di2018safety} addresses obstacle avoidance within task priority framework for a selected control points by computing the minimum obstacle distance at 30 Hz from depth image by exhaustively scanning the neighbourhood of the control points until it finds the nearest obstacle pixel. 



Apart from collision checking and minimum distance computation, online motion control also relies on a minimization problem with different task constraints to select optimal control inputs in real time. Motion control strategies with online collision avoidance are usually handled by exploiting the task priority framework in which null-space projection matrices are employed \cite{b1}, \cite{b2} and \cite{b6}. This approach applies the null space of higher priority task to achieve lower priority tasks. For this, a null space projector matrix is derived using pseudo-inverse. To facilitate task insertion or removal, an activation matrix of dimension equal to the dimension given task is used. The problem with this approach is that the use of activation matrix resulted in what is commonly known as a practical discontinuity in which a mathematically continuous rapidly varying property of the control becomes discontinuity when implemented on actual robot. The work in \cite{b2} proposed a task oriented regularization to tackle the problem of practical discontinuity i.e. it handles any general number of equality and inequality tasks while allowing task activation and deactivation without practical discontinuity.

\section{General Framework for Collision Avoidance}\label{sec:Method}
%
%
In the proposed collision avoidance algorithm, the environment and the robot are described inside two separate but equal discretized regular grids of cubic voxels, with grid resolution equals to the vovel size. At every time step, both the robot and the environment maps are updated by defining the occupied voxels in their respective grids and the distance transform of the environment map is computed in real time. From this distance map, we can efficiently extract obstacle distance for few key points on the robot to activate collision avoidance tasks. The selection of these key points, also known as control points, is usually located at the center of the spheres that approximate the robotic links.

With this general approach in mind, there are challenges associated to exploiting GPU for the collision avoidance algorithm. The first, perhaps the starting point for our work is to separate various components of the collision avoidance algorithm based on whether they are suitable for parallelization or not. The collision avoidance algorithm based on a task-priority controller (Alg.~\ref{algo:self}) is implemented to run on a CPU due to their nature, while the EDT (Alg.~\ref{alg:EDT}) benefits from the massive parellization available in GPU. Secondly, both parts of the algorithm  exchange information which may lead to bandwidth bottleneck in copying data from CPU to GPU and vice versa and therefore minimal data copying procedure is intended. Here, 3D pointcloud is down-sampled before it is copied to GPU memory.  After computing the EDT, copying back the entire Distance map to CPU  is avoided by simplifying the robot model so that only minimum obstacle distance to selected points suffice for the collision avoidance algorithm. Lastly, as each voxels are associated to a memory in the GPU, large map size will result in a significant memory consumption. As our development targets only online collision avoidance, a map size that accommodate the immediate workspace of the robot is initialized thereby leveraging memory usage. 

To efficiently exploit the redundancy in the manipulator, the obstacle avoidance task is activated only for those parts of the robot whose distance to the closest obstacle is less than the radius of the spheres used to enclose, i.e. discretize, the robot structure. The discontinuity due to task activation and deactivation is mitigated by exploiting the task priority framework and the task oriented regularization, thereby imposing smooth behavior in the joint control velocity. 

\begin{figure*}[t]
 \centering
\includegraphics[width=0.9\linewidth]{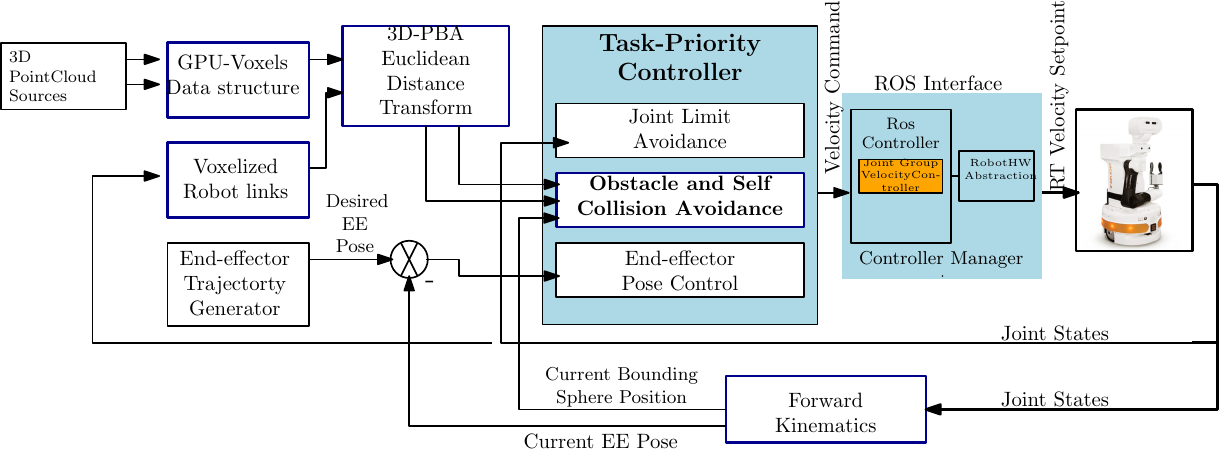}
  \caption{The Proposed Collision Avoidance control scheme. The main components of the  proposed system  are embedded in the blocks marked with blue. The GPU-Voxels data structure updates and stores occupancy information of the environment from live 3d point cloud sources(\ref{subsec:obstacle_rep}). For self Collision avoidance, the robot's link is converted to voxels and stored in a separate GPU-voxels map. Based on the occupancy information of the environment and the robot, the 3D-Euclidean distance is computed at very high rate using the PBA algorithm\ref{exact EDT}. This is utilized during the Obstacle and Self-collision Avoidance of the Task-Priority Algorithm described in Section~\ref{obstacle avoidance}. The ROS control Hardware Abstraction in the final block handles out of the box interface with our Task Priority Control in real time (RT).  }\label{fig:sys_block}
 \vspace*{-3mm}
\end{figure*}

\subsection{Obstacle Representation}
\label{subsec:obstacle_rep}
\begin{figure}[t]
 \centering
 \subfigure{
  \includegraphics[height=0.13\textwidth]{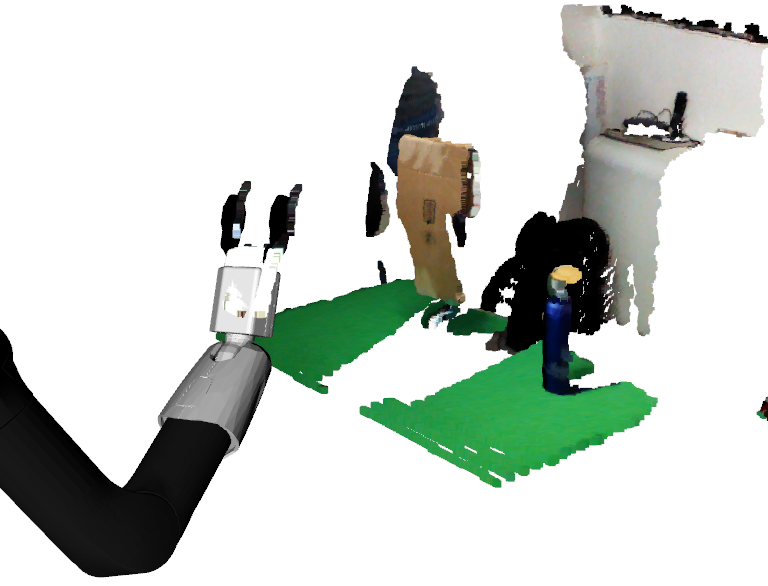}
 }
  \subfigure{
  \includegraphics[height=0.13\textwidth]{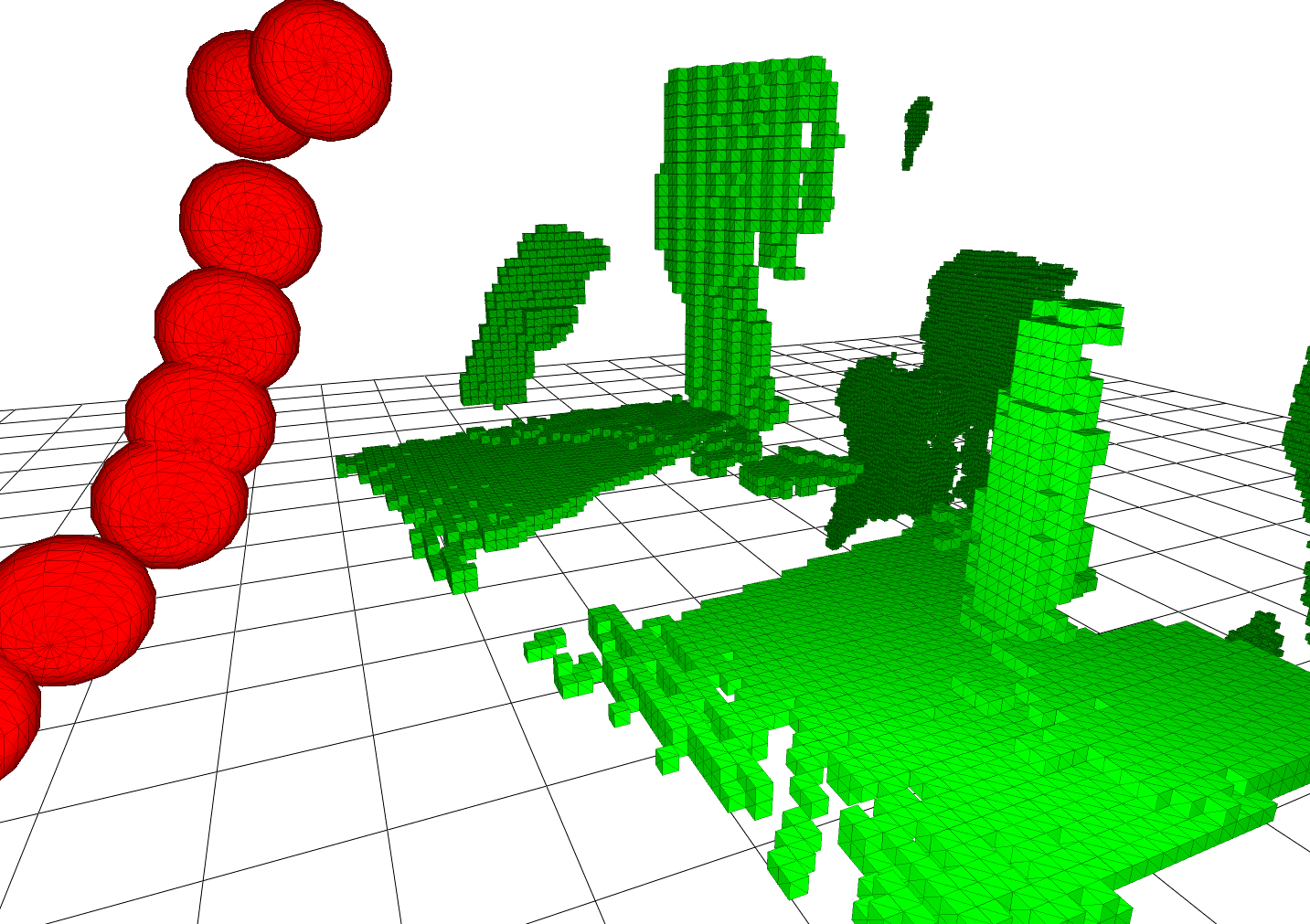}
 }
 \caption{ Raw point cloud processed to remove noise and robot part from the scene: Raw point cloud visualization in Rviz containing points coming from the robot gripper (left) and GPU voxels representation (right).
 \label{fig:raw_vs_filtered}}
 \vspace*{-3mm}
\end{figure}

In our proposed approach the environment (obstacles) is represented as a GPU voxels occupancy grid \cite{b7} in a probabilistic way. To create this environment map, the acquired point clouds are raw-copied into GPU memory to perform a statistical outlier filtering based on their Euclidean distance to their neighbors and followed by a transformation to a fixed coordinate. 
After transformation, the coordinates of each point are discretized to determine the according Voxel, whose occupancy status is updated as a Bayesian process.  During insertion of points into the environment map corresponding Voxels’ meanings in the robot map are reviewed if they are occupied by the robot model to prevent the insertion of  points originating from robot parts, see  Fig.~\ref{fig:raw_vs_filtered}. The impact of occlusions can also be tackled using multiple source of point cloud, without affecting the speed of the algorithm. 

For the self-collision avoidance task, the mesh model of the robot is rasterized into a binary 3D voxel grid to create a binary voxel model of the robot using an offline software tools such as in \cite{b8}. This offline generated voxel representation of the robot's link will be inserted into a map according to the joint state information of the actual robot to represent obstacles for the self-collision avoidance task as shown in Fig.~\ref{fig:rob_env} and Fig~\ref{fig:centers}. By inserting the voxelized shape of each corresponding robot link, a full discrete representation of the robot is added into a robot map with same dimension as the environment map.

\subsection{Robot Representation} \label{subsec:robot_repre}
The robot model is critical when performing minimum distance computation as complex models such as triangle meshes consume considerable amount of time while an over simplified models might lead to collision. To speedup distance query, various robot model approximation schemes has been proposed. The predominant approach utilizes simple primitive shapes in a hierarchical manner~\cite{b9,b10, C1}. An alternative approach to the hierarchical representation is to utilize complex geometric primitives such as cylinders, boxes and capsules~\cite{C14,safeea2019line,safeea2019minimum}. These geometries minimize the number models involved (one of these shapes can in-close an entire robot link). However, additional computations are required to determine points of minimal distance on both the models and the environment side in addition to multiple distance queries. Even more, these computations are not tailored toward parallel operation.
 
In this work, we propose a collision model composed of minimal number of  spherical volumes with different radii across different links that discretize each link with low approximation error and without incurring additional computation on both CPU and GPU side. Here, spheres are merely used for enclosing  the robot geometry and  we do not rely on the hierarchy of spheres for distance query as opposed to the approaches reported in \cite{b9,b10, C1}.
To generate this set of bounding spheres, first an oriented bounding box (OBB) of each link in the kinematic chain is computed. Each OBB is represented by a center point, an orientation matrix and three half-edge lengths and are computed only once and in offline, therefore no overhead is added in run-time. 
The smallest diagonal dimension on the face of the OBB will be assigned as the diameter of the spheres and are placed along the longest dimension of the box. For an OBB with dimensions $d_1$, $d_2$, $d_3$ and $d_1 \leq d_2 \leq d_3$, spheres of diameter $\sqrt{d^2_1 +d^2_2}$ where  will be utilized.  This guarantees inscription of the OBB within the sequence of $\ceil[\big]{\frac{d_3}{\sqrt{d^2_1 +d^2_2}}+1}$ spheres, thereby allowing finite minimal number on each link. 
The distance queries can easily be compared to the radius of each sphere for collision detection thanks to their rotational invariant property.
\subsection{Exact EDT Computation}
\label{exact EDT}
 Given a 3D voxel grid of $\mathbb{G} = n \times n \times n$ voxels,  the EDT problem is to determine the closest occupied voxel for each voxel in the grid.
 This problem is closely related to the Voronoi diagram computation, i.e. The EDT of a binary grid  can be thought of as a discretized version of the Voronoi
diagram whose Voronoi sites are the occupied voxels of the grid.

The parallel banding  algorithm (PBA) proposed in \cite{b99} computes exact EDT on GPU by computing partial Voronoi which involves three phases where each of them are parallelized using bands to increase the number of threads (Alg.~\ref{alg:EDT}).  For a 3D voxel grid  $\mathbb{G}$ of size $n$, Consider a slice $\mathcal{I}_k$ at $z=k$ where $k$ ranges from $k=0$ to $n-1$. This slice is basically a 2D binary image of $n \times n$ size. Algorithm~\ref{alg:EDT} is a high level pseudo code summarizing the three steps as follow: 
\subsubsection{Step-1 (BandSweep (Line 3-4)):} Calculates $S_{i,j,k}$ i.e, the nearest Voronoi site, among all sites in slice $k$ and row $j$, of the voxel $(i, j, k)$. To efficiently employ threads, the slice $\mathcal{I}_k$ is divided into m1 vertical bands of equal size, and use one thread to handle one row in each band, performing the left-right sweeps followed by across band propagation. $n*m_1$ threads are utilized per slice in this step.
\subsubsection{Step-2 (ComputeProximateSite  (line 6)):} i.e, $\mathcal{P}_i$. Let $S_i = {S_{i,j,k} \mid S_{i,j,k} \neq \emptyset, j =
0, 1, 2, . . . , n-1}$ be the collection of  closest sites for all voxels in column i of slice k.
The sequential implementation to determine $\mathcal{P}_i$ is to sweep sites in $S_i$ from topmost to bottommost, while maintaining a stack of sites that are potentially proximate sites, i.e; sites whose voronoi region intersects with column i. For every new site c, we evaluate whether the site at the top of the stack is dominated by c and the site a at the second top position in the stack. If so, b is popped out of the stack, and  c is pushed onto the stack, while the sweeping continue until the column i is completed by returning the stack containing $\mathcal{P}_i$.
To parallelize this sequential operation, each column $S_i$ is partitioned horizontally into $m_2$ bands, thereby employing $n*m_2$ gpu threads per slice to perform the above computation.

\subsubsection{Step-3 (QueryNearestSite (line 7))}
This step utilizes $\mathcal{P}_i$ from step 2 to compute the closest site for each voxel V in column i of slice k top-down by checking  two consecutive sites a and b in $\mathcal{P}_i$ in increasing y coordinate. If a is closer, then a is the closest site to V, if not a will be removed and the process continues with b and the next site. This is also performed in $m_3$ horizontal bands; therefore $n*m_3$ threads deployed per slice. 
\subsubsection{Step-4 (Extention to 3D (line 8-11))}
At the end step-3, the voxel map contains a stack of 2D Voronoi diagrams. To extend this to 3D, step-2 and step-3 are repeated for all columns in the direction of the stack (line 8-11).  

\begin{algorithm}[t]
	\DontPrintSemicolon 
	\KwIn{$\mathbb{G}: [0..n-1]\times [0..n-1]\times [0..n-1] \rightarrow \{0,1\}$, $m_1$, $m_2$, $m_3$}
	\KwOut{$EDT(\mathbb{G})$}
	\SetKwRepeat{Do}{do}{while}
    
    \For {$k =0$  to $n-1$}{
    
    	$\mathcal{I}_k \leftarrow$ Slice($\mathbb{G}, z=k$)\;
    	$S^{'}_{i,j,k} \leftarrow$ \text{BandSweep}($\mathcal{I}_k, band=m_1$)    	
    	\label{line:phase_bandsweeping}\;
    
    $ S_{i,j,k}  \leftarrow \text{PropagateAcrossBand}(S^{'}_{i,j,k}) $\;
    \For {$i \in [0..n-1]$}
    { 
        $ \mathcal{P}_i \leftarrow             \text{ComputeProximateSite}(S_i, band=m_2 )$ \label{line:computeproximate2d}\;
        $ EDT _\text{i,j,.} \leftarrow             \text{QueryNearestSite}(\mathcal{P}_i, band=m_3)$
    }
    }
     \For {$(i,j) \in [0..n-1]\times [0..n-1]$}
    { 
        \For {$k =0$  to $n-1$}
    { 
       $ \mathcal{P} _{i,j,.} \leftarrow             \text{ComputeProximateSite}(S_k, band=m_2 )$ \label{line:computeproximate}\;
        $ EDT _\text{i,j,k} \leftarrow             \text{QueryNearestSite}(\mathcal{P}_{i,j, .}, m_3)$ \label{line:querysite}\;
       }
    }
	\caption{Parallel\_Banding\_EDT }\label{alg:EDT}
\end{algorithm}

\subsection{Obstacle Avoidance Task}
\label{obstacle avoidance}
The obstacle avoidance control is  formulated as an $m$ dimensional kinematic constraint associated to $m$ bounding spheres distributed along the links of the robot.  Consider, a collision avoidance task vector $X_c \in {\mathbb{R}}^m$, with $i$-th element $x_{c_i}$ corresponds to a scalar task associated to $i$-th sphere on the robot link and defined as the  distance between it's center $\mathbf{C_i}$ and it's nearest obstacle voxel coordinate $\mathbf{O_i}$:
 \begin{equation}
 \label{dist}
  x_{c_i} =\left\lVert\mathbf{O_i}-\mathbf{C_i}\right\rVert_2
 \end{equation}
 To implement this, we keep track of  the coordinates of voxels at the center of the spheres described in section~\ref{subsec:robot_repre} (line 10) and their closest obstacle  coordinate (line 11) of Alg.~\ref{algo:self}. 
 The goal of collision avoidance control is to keep the task variable $x_{c_i}$ bigger than the radius of the $i$-th bounding sphere, $x_{M_i}$.
 \begin{equation}
 \label{ref}
  x_{c_i} \geq x_{M_i}
 \end{equation}
The task Jacobian $J_i$, for the $i$-th scalar task $x_{c_i}$ defines the direction that pushes the particular  bounding sphere away from the nearest obstacle. This Jacobian is given by single row matrix derived by projecting position Jacobian $J_{c_i}$ at $\mathbf{C_i}$ in the vector direction connecting $\mathbf{C_i}$ and $\mathbf{O_i}$.
 \begin{equation}
 \label{jac}
  J_{i} = \left(-\frac{\mathbf{O_i}-\mathbf{C_i}}{x_{ci}}\right)^T J_{c_i}(\textbf{q})
 \end{equation}
 At every control cycle the task constraint eq.~\eqref{dist} and task Jacobian eq.~\eqref{jac} are updated in Alg.~\ref{algo:update_jac}, (lines 2-6).
 The linear velocity in the opposite direction w.r.t the closest obstacle at $\mathbf{C_i}$ is related to the joint velocity through the  differential kinematic equation $\Dot{x}_i =  J_{i}\, \Dot{\textbf{q}}$.
Subsequently the suitable reference velocity for collision avoidance can be defined to be proportional to the difference between current task variable $x_{c_i}$ and any $x^* > x_{M_i}$.
Therefore, the velocity that drives $x_{c_i}$ toward its corresponding objective $x^*$ is
 \begin{equation}
 \label{ref_vel}
  \Dot{\bar{x}}_{c_i} = \kappa(x^*-x_{c_i})
 \end{equation}
 To generalize the above scalar representation of the collision avoidance task in to an m-dimensional task, we make use of the following notation: $X_c \in \mathbb{R}^m$ is the task variable vector, $\dot{\bar{X}}_c \in \mathbb{R}^m$ is a vector of suitable reference rate for the task vector $X_c$ and $J_{CA}\in \mathbb{R}^{m\times n}$ is the obstacle avoidance task Jacobian. They can be described using the scalar quantities shown in eqs.~\eqref{dist} - \eqref{ref_vel} as follows:
 \begin{equation}
\begin{split}
 X_c &=\begin{bmatrix}
 x_{c_1}\
 x_{c_2}\ 
 \cdots\
 x_{c_i}\
 \cdots\
 x_{c_m}
\end{bmatrix}^T ; \\
\dot{\bar{X}}_c&=\begin{bmatrix}
 \Dot{\bar{x}}_{c_1}\
 \Dot{\bar{x}}_{c_2}\ 
 \cdots\
 \Dot{\bar{x}}_{c_i}\
 \cdots\
 \Dot{\bar{x}}_{c_m}
\end{bmatrix}^T
\end{split}
 \end{equation}
 and the collision avoidance task Jacobian is given by
 \begin{equation}
 J_{CA} = \begin{bmatrix}
 J_{1} \
 J_{2} \
 \cdots \
 J_{i} \
 \cdots\  
 J_{m}
\end{bmatrix}^T
 \end{equation}
The main call  to the collision avoidance starts in Alg.~\ref{algo:self} by initializing the maps and loading bounding sphere centers in link frame (line 1). At every control cycle, new point cloud and the voxelized robot links are updated and inserted into their corresponding voxel map while the position of the sphere centers are computed  using forward kinematics (line 2-8). This is followed by a call to a gpu kernel that implements Alg.~\ref{alg:EDT} to compute the EDT of the maps (line 9). Note that, the EDT transform is stored and updated in GPU memory while only the minimum distance obstacle coordinate to the sphere centers $\{\mathbf{C_1} \dots \mathbf{C_m}\}$ are copied back to the CPU to update the Jacobian (eq.~\ref{jac}) and control (eq.~\ref{ref_vel}) by calling Alg.~\ref{algo:update_jac} in line 12 of Alg.~\ref{algo:self}.
 \subsection{Self-Collision Avoidance}
\begin{figure}
 \centering
 \includegraphics[width=15cm, height=5cm,
  keepaspectratio]{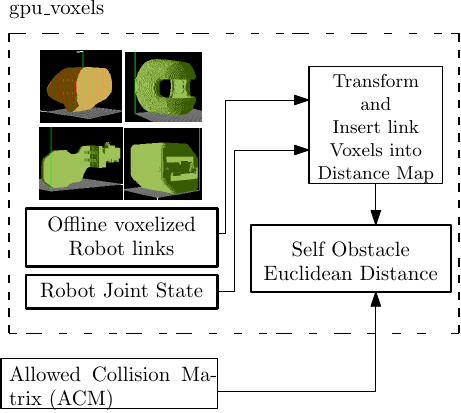}
 \caption{Pipeline for Self collision minimum distance computation: The voxel model of Links that are not in the ACM are transformed and inserted inside a GPU\_voxels map followed by EDT. }
 \label{fig:self_pipeline}
\end{figure}

Self-collision avoidance of an arbitrary link utilizes EDT of a set of prior links to determine the closest points. In general, the obstacle of a kinematic chain in self-collision sense is composed of all other links in the robot body $L_1$, $L_2$, $\hdots$, $L_{i}$ that are not in the Allowed Collision Matrix (ACM), see Fig.~\ref{fig:self_pipeline}, and it is given as:
\begin{equation}
 O_{L} =\{L_1, L_2, \hdots, L_{i}\}
 \label{eq:selfclink}
\end{equation}
All links in the set $O_L$ are voxelized offline,  transformed and inserted into voxel grid according to their corresponding joint state to compute the EDT for self-collision avoidance, as in Alg.~\ref{algo:self}, lines 3-6. For our particular robotic platform, i.e Tiago robot, links shown in yellow voxels in Fig.~\ref{fig:centers} form the set of all possible self collision objects for the arm kinematic chain indicated by the red spherical balls.  


Once the closest points are computed, for instance, $o_{s_i}$ for the a task sphere of $c_i$, a self-collision avoidance task restricts the relative motion of the two points in the direction of the line connecting them.
In a similar way to obstacle avoidance, the $i$-th scalar self-collision avoidance task $x_{s_i}$ and It's corresponding Jacobian $J_{s_i}$ are given as:
\begin{equation}
 x_{s_i} =\left\lVert\mathbf{O_{s_i}}-\mathbf{C_i}\right\rVert _2
 \label{eq:selftaskvar}
\end{equation}
and 
\begin{equation}
  J_{s_i} = \left(-\frac{\mathbf{O_{s_i}}-\mathbf{C_i}}{x_{si}}\right)^T J_{c_i}(\textbf{q})
  \label{eq:selftaskjac}
\end{equation}
\begin{algorithm}[t]
 \caption{Obstacle and Self\_Collision\_Avoidance}
 \label{algo:self}
 \begin{algorithmic}[1]
 \renewcommand{\algorithmicrequire}{\textbf{Input:}}
 \REQUIRE 
 {\begin{itemize}
 \item[] Bounding Sphere Centers in Link frame \{$\mathbf{C_1}$\dots$ \mathbf{C_m}$ \}
 \item[] Bounding Sphere Radius \{$x_{M_1}$\dots$x_{M_m}$ \}
 \end{itemize}%
 } 
 \STATE Initialisation: \texttt{GPU\_Voxels} Self Collision Obstacle Map $\mathbb{M_S}$ and Environment Obstacle Map $\mathbb{M_I}$
 \WHILE{True}
 \STATE update Joint\_States \textbf{q} =\{$q_1$, $q_2$,\dots $q_n$\}
 \STATE \{$\mathbb{M_S}$, $\mathbb{M_I}$\} $\gets$ ClearMap
 \STATE $ O_{L_k} \gets$ UpdateLinks($\textbf{q}$)
 \STATE $\mathbb{M_S} \gets$ Insert($O_{L_k}$)
 \STATE $\mathbb{M_I} \gets$ Insert(\texttt{Point\_Cloud}) 
 \STATE $ \{\mathbf{C_1}$\dots$ \mathbf{C_m}\}_\text{global\_frame} \gets \text{ForwardKinematics}(\textbf{q})$
 \STATE \{$\mathbb{M_S}$, $\mathbb{M_I}$\} $\gets$ PBA\_EDT($\mathbb{M_S}$, $\mathbb{M_I}$, $m_1$, $m_2$, $m_3$)
 \STATE \begin{varwidth}{\linewidth}
  $ \{\mathbf{O_{s_i}} \dots \mathbf{O_{s_m}}; \mathbf{O_i} \dots \mathbf{O_m}\}_\text{global\_frame}$~$\gets$ \par
  \hskip\algorithmicindent 
  ExtractClosestObstaclePose$(\mathbb{M_S}, \mathbb{M_I}, \{\mathbf{C_1} \dots \mathbf{C_m}\}) $
  \end{varwidth}
 \STATE \begin{varwidth}{\linewidth}
  $X_c $, $\dot{\bar{X}}_c$, $ J_{CA}$, $ J_{SA}$, $A$ ~ $\gets$ \par
  \hskip\algorithmicindent 
  Update\_Col\_Jac$(\mathbf{C_1}$\dots$ \mathbf{C_m}; \mathbf{O_{s_i}} \dots \mathbf{O_{s_m}}; \mathbf{O_i} \dots \mathbf{O_m}) $
  \end{varwidth}
 \STATE $\dot{q}_{p}$ $\gets$ UpdatePriorityLevel($A, X_c , \dot{\bar{X}}_c, J_{CA}, J_{SA}, \dot{q}_{p-1}$)
 \ENDWHILE
 \end{algorithmic} 
 \end{algorithm}
\begin{algorithm}[t]
 \caption{Update\_Col\_Jac}
 \label{algo:update_jac}
 \begin{algorithmic}[1]
 \renewcommand{\algorithmicrequire}{\textbf{Input:}}
 \REQUIRE 
 {\begin{itemize}
 \item[] $ \{\mathbf{C_1}$\dots$ \mathbf{C_m}; \mathbf{O_{s_i}} \dots \mathbf{O_{s_m}}; \mathbf{O_i} \dots \mathbf{O_m}\}_\text{global\_frame}$
 \end{itemize}%
 } 

 \FOR {$i = 1$ to $m$}
 \STATE $J_{c_i}(\textbf{q})$ $\gets$ Compute Position Jacobian at $\mathbf{C_i}$
 \STATE $x_{s_i}$ $\gets$ $\left\lVert\mathbf{O_{s_i}}-\mathbf{C_i}\right\rVert_2$
 \STATE $x_{c_i}$ $\gets$ $\left\lVert\mathbf{O_i}-\mathbf{C_i}\right\rVert_2$
 \STATE \begin{varwidth}{\linewidth}
 Compute Self Collision Avoidance Jacobian at $\mathbf{C_i}$ \par
  \hskip\algorithmicindent
 $J_{s_i}(\textbf{q})$ $\gets$ $(-\frac{\mathbf{O_{s_i}}-\mathbf{C_i}}{x_{s_i}})^T J_{c_i}(\hat{q})$
 \end{varwidth}
 \STATE \begin{varwidth}{\linewidth}
 Compute Obstacle Avoidance Jacobian at $\mathbf{C_i}$ \par
  \hskip\algorithmicindent
 $J_{i}(\textbf{q})$ $\gets$ $(-\frac{\mathbf{O_i}-\mathbf{C_i}}{x_{ci}})^T J_{c_i}(\textbf{q})$
 \end{varwidth}
 \STATE \begin{varwidth}{\linewidth}
 Compute Activation Value $A_{s_{(i,i)}}$, $A_{o_{(i,i)}}$ \par
  \hskip\algorithmicindent
  $A_{s_{(i,i)}}$ $\gets$ Sigmoid$(x_{s_i}, x_{M_i}, b_i)$ \par
  \hskip\algorithmicindent
  $A_{o_{(i,i)}}$ $\gets$ Sigmoid$(x_{c_i}, x_{M_i}, b_i)$
 \end{varwidth}
 \STATE \begin{varwidth}{\linewidth}
 Compute Control $\Dot{\bar{x}}_{s_i}$, $\Dot{\bar{x}}_{c_i}$ \par
  \hskip\algorithmicindent
  $\Dot{\bar{x}}_{s_i}$ $\gets$ $\kappa(x^*-x_{s_i})$ \par
  \hskip\algorithmicindent
  $\Dot{\bar{x}}_{c_i}$ $\gets$ $\kappa(x^*-x_{c_i})$
 \end{varwidth}
 \ENDFOR
 \STATE \begin{varwidth}{\linewidth}
 \vspace*{1mm}
 Return
 \hskip\algorithmicindent
 $X_c $, $\dot{\bar{X}}_c$, $ J_{CA}$, $ J_{SA}$, $A$
 \end{varwidth}
 \end{algorithmic} 
 \end{algorithm}

\subsection{Task Priority Controller Formulation}

For a general robotic system with n-DoF, in a given configuration, $\mathbf{q} =[q_1 \ q_2\ \cdots\ q_n]^T$, the forward kinematics of a particular task $\mathbf{x}\in \mathbb{R}^m$ can be expressed as a function of joint configuration:
\begin{equation}
\mathbf{x(t)= x(\textbf{q}(t))}
\end{equation}
For such task variable, we also assume the existence of Jacobian relationship between task space velocity $\Dot{x}$ and the joint velocity vector $\Dot{\textbf{q}}$ as
\begin{equation}
\mathbf{\Dot{x} = J(q)\Dot{q}}
\end{equation}
where $J(\textbf{q}) \in \mathbb{R}^{m \times n}$ is the Jacobian matrix. 
Given a reference task space velocity vector $\Dot{\overline{x}}$, joint velocity vector $\Dot{\textbf{q}}$ that satisfies $\Dot{\overline{x}}$ in the least-square sense, can be computed using Pseudo inverse as
\begin{equation}
\label{minimize1}
\min_{\Dot{\textbf{q}}}{\left\lVert \Dot{\overline{x}}-J \Dot{\textbf{q}} \right\rVert}^2\implies \Dot{\textbf{q}}=(J^T J)^\# J^T \dot{\overline{x}}
\end{equation}
The manifold of all solutions can be derived by introducing null space projector that ensures task velocity remains unchanged:
\begin{equation}
\Dot{\textbf{q}} = (\textbf{J}^T\textbf{J})^\# \textbf{J}^T \dot{\overline{\textbf{x}}}+(\textbf{I}-(\textbf{J}^T\textbf{J})^\# \textbf{J}^T \textbf{J})\dot{\textbf{q}}_0
\end{equation}
where $(.)^\#$ represents generalized pseudo-inverse and $\dot{q}_0$ denote a joint velocity that produce an orthogonal component of $\Dot{\overline{x}}$ i.e. not affecting the desired task. 

Task insertion and removal is handled using a diagonal task activation matrix $A \in \mathbb{R}^{m \times m}$ associated with each task $x \in \mathbb{R}^m$ whose value is given by 

\begin{equation}
A_{(i,i)} = \begin{cases}
1 &\text{$x_i \geq x_{M_i}$ (Activated)}\\
s_i(x_i) &\text{$x_{Mi}-b_i \leq x_i \leq x_{Mi}$ (transition)}\\
0 &\text{$x_i \leq x_{Mi}-b_i$ (Deactivated)}\\
\end{cases}
\end{equation}
where $s_{i}(x_i)$ is the sigmoid function given by
\begin{equation}
s_{i}(x_i) = \dfrac{1}{2}\left(\cos\dfrac{(x_i-x_{Mi})\pi}{b_i}+ 1\right)
\end{equation}

where $b_i$, $x_i$ and $x_{M_i}$ are the $i$-th task transition buffer ( a nonzero parameter defined for tasks with inequality constraints and their values are set by the user depending on the type of the task. For collision avoidance task, 0.02m utilized.), $i$-th elements of the task $x$ and the task activation threshold of the $i$-th task respectively. with the activation matrix $A$, the minimization problem in eq.~\eqref{minimize1} takes the form $\min_{\Dot{q}}{\mathbf{\left\lVert A(\Dot{\overline{x}}-J \Dot{q}) \right\rVert}^2}$ ensures that only tasks that are active to be considered in the problem by multiplying the  terms with the activation matrix of the task, thereby guaranteeing inactive tasks not to consume control directions, effectively increasing arbitrariness space of the solution.

To avoid large joint velocities due to task singularity, singular value oriented regularization term is required to penalize joint velocities in the task singularity direction. Even more, to tackle practical discontinuity due to task activation and deactivation, a task oriented regularization term is added to selectively penalize scalar tasks that are in transition phase. Thus, introducing both singular value oriented and task-oriented regularization in to the minimization problem in \eqref{minimize1} gives 
\begin{equation}
\label{task_oriented}
\min_{\Dot{q}}{\mathbf{\left\lVert A(\Dot{\overline{x}}-J \Dot{q}) \right\rVert}^2 + \mathbf{\left\lVert J \Dot{q} \right\rVert_{A(I-A)}}^2 +\mathbf{\left\lVert V^{T} \Dot{q} \right\rVert_{P}}^2}
\end{equation}
where $V^{T}$ is the right orthonormal
 matrix of the SVD decomposition of $J^T A J= U \Sigma V^T $ and
 $P$ is a diagonal regularization matrix where each diagonal element $p_{(i,i)}$ is a bell-shaped function of the corresponding singular value of $J$, or zero if the corresponding singular value do not exist and the notation $\|\cdot\|_P$ indicates the weighted norm, i.e. $\| \dot{q} \|^2_{P} = \dot q^T P \dot{q}$.
The generalized solution of \eqref{task_oriented} is then given by
\begin{equation}
\label{task_oriented_solution}
\begin{split}
\Dot{\textbf{q}} &= (\textbf{J}^T\textbf{AJ}+\textbf{V}^T\textbf{PV})^\# \textbf{J}^T \textbf{A A} \dot{\overline{\textbf{x}}}
\\
& +(\textbf{I}-(\textbf{J}^T\textbf{AJ}+\textbf{V}^T\textbf{PV})^\# \textbf{J}^T\textbf{AA} \textbf{J})\dot{\textbf{q}}_{0}
\end{split}
\end{equation}
where $\dot{q}_{0}$ can be used to perform lower priority tasks in hierarchy. 

\section{Experiments} \label{sec:experiments}
\subsection{Experimental Setup}
The 7-DoF Tiago robot arm is used in the experiments to interact with a dynamic obstacle. The scene is captured at 30$\,$Hz by an ASUS Xtion 3D camera available on the Tiago robot head and eventually transformed to the base frame of the robot using the approach described in section \ref{subsec:obstacle_rep}. 
An external computer hosting an NVIDIA GeForce GPU (GeForce GTX 1080 Ti) with 3584 CUDA cores for parallel distance computation is used for this purpose. Even though the robotic arm can only span a volume of less than 1.2 m $\times$ 1.2 m $\times$ 1.8 m, the experiments are performed on a 1.92 m $\times$ 1.92 m $\times$ 1.92 m physical volume
and evaluated the real time property of maps as large as 20.48 m $\times$ 20.48 m $\times$ 5.12 m, i.e (512$\times$512$\times$128 with 4$\,$cm voxel size).
During all experiment scenarios, the following task hierarchy is considered: (1) Joint limiter (2) Collision  Avoidance and (3) End-effector Pose control, where the priority goes from the highest to the lowest with the joint limiter and collision avoidance tasks are activated and deactivated according to an activation matrix. By leveraging ROS, communication between our application, 3d camera and joint actuators are handled seamlessly. However, the PBA implementation is done in CUDA C, therefore requiring additional layer to allow ROS type data communication. This is particularly apparent when copying back nearest obstacle information from GPU memory to our application.

\begin{table}[b]
\caption{GPU Voxels Exact Euclidean Distance Computation Time with Voxel size of 0.5cm, 1cm and 2cm. Throughout the experiment the  CPU runs at about 500 Hz without including data copying operation}
\begin{tabular*}{\hsize}{@{\extracolsep{\fill}}lllll@{}}
\toprule[1pt]\midrule[0.3pt]
\makecell{Map \\ Dimension\\($[Vox]^3$)} & \makecell{No. of \\ Occupied \\Voxels} & \makecell{Clear \\ Map \\ (ms)} & \makecell{Insert \\ PointCloud \\(ms)} & \makecell{Compute \\ Distance \\PBA (ms)}\\
\toprule
\toprule
Voxel size 2cm & & & & \\
\toprule 
192$\times$192$\times$128 & 6214 & 0.2284 & 0.3281 & 1.7601\\
256$\times$256$\times$128 & 9080 & 0.4565 & 0.3179 & 3.2475\\
512$\times$512$\times$128 & 12274 & 0.8268 & 0.2580 & 8.1972\\
\toprule 
Voxel size 1cm & & & & \\
\toprule 
192$\times$192$\times$128 & 6863 & 0.9978 & 1.0639 & 2.4013\\
256$\times$256$\times$128 & 10014 & 1.2085 & 0.9460 & 3.9767\\
512$\times$512$\times$128 & 18274 & 1.3446 & 0.9426 & 11.5601\\
\toprule
Voxel size 0.5cm & & & & \\
\toprule
192$\times$192$\times$128 & 7263 & 1.0478 & 1.2639 & 2.8013\\
256$\times$256$\times$128 & 12014 & 1.2985 & 0.9962 & 4.2767\\
512$\times$512$\times$128 & 19304 & 1.5446 & 1.1426 & 12.0601\\
\toprule
\label{pipeline time}
\end{tabular*}
\vspace*{-3mm}
\end{table}

\subsection{Dynamic Obstacle Avoidance Results}
\begin{figure}[t]
 \centering
 \subfigure{
  \includegraphics[width=0.129\textwidth]{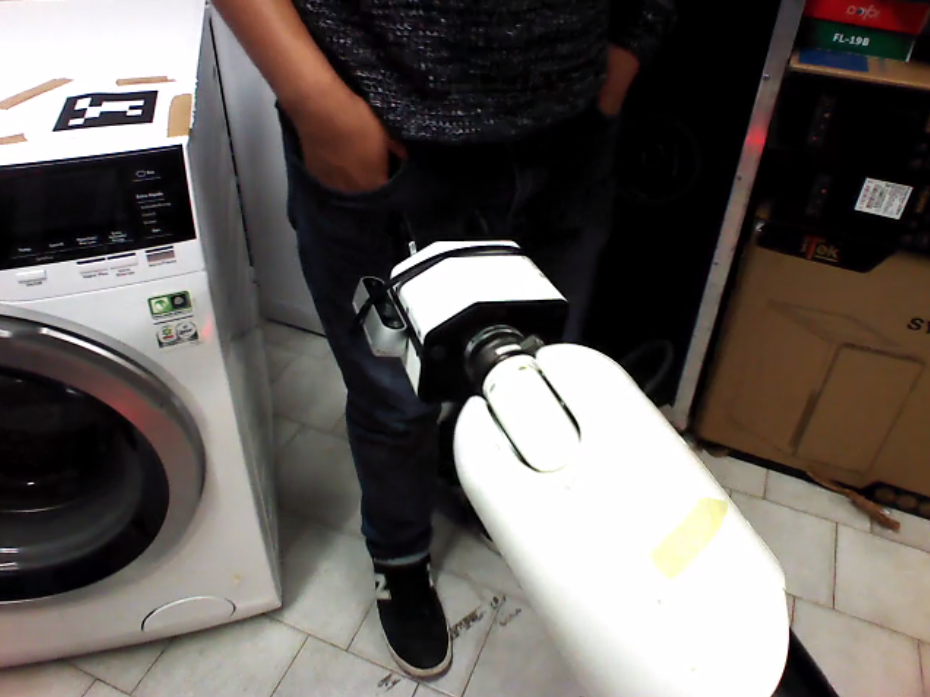}
 }
  \subfigure{
  \includegraphics[width=0.129\textwidth]{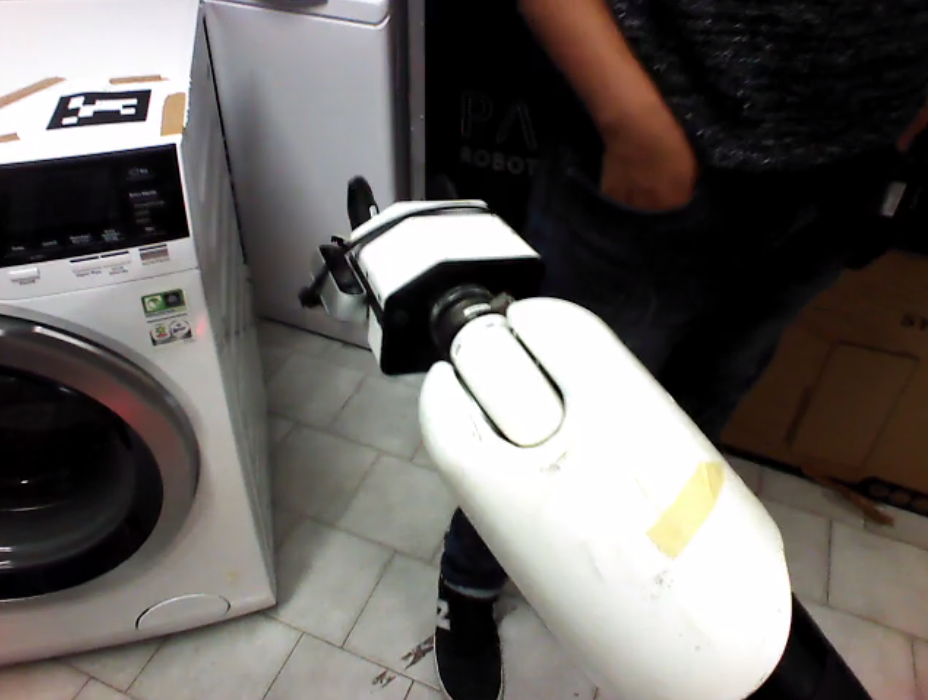}
 }
  \subfigure{
  \includegraphics[width=0.129\textwidth]{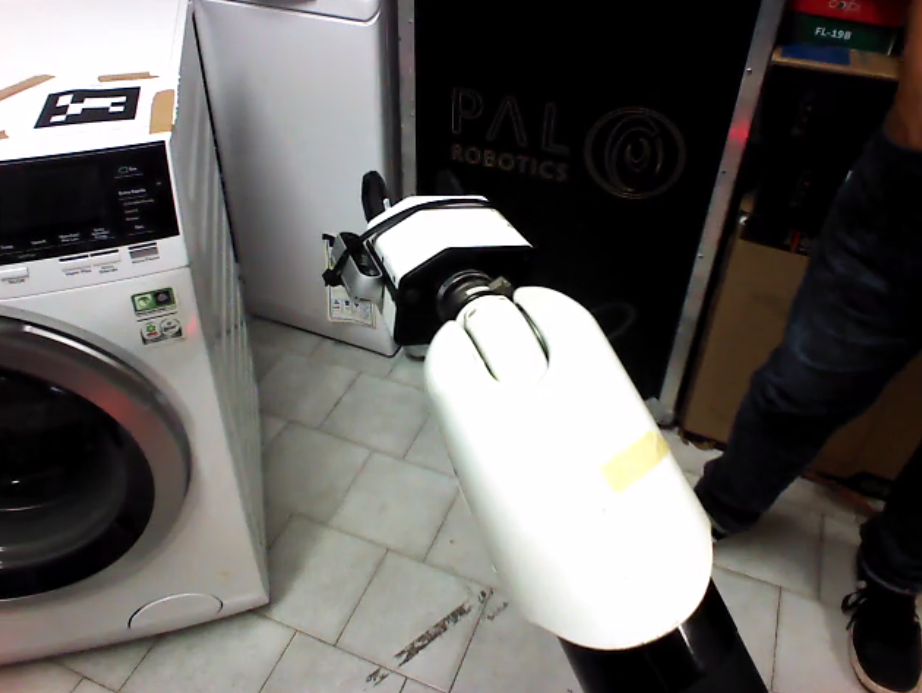}
 }
  \subfigure{
  \includegraphics[width=0.129\textwidth]{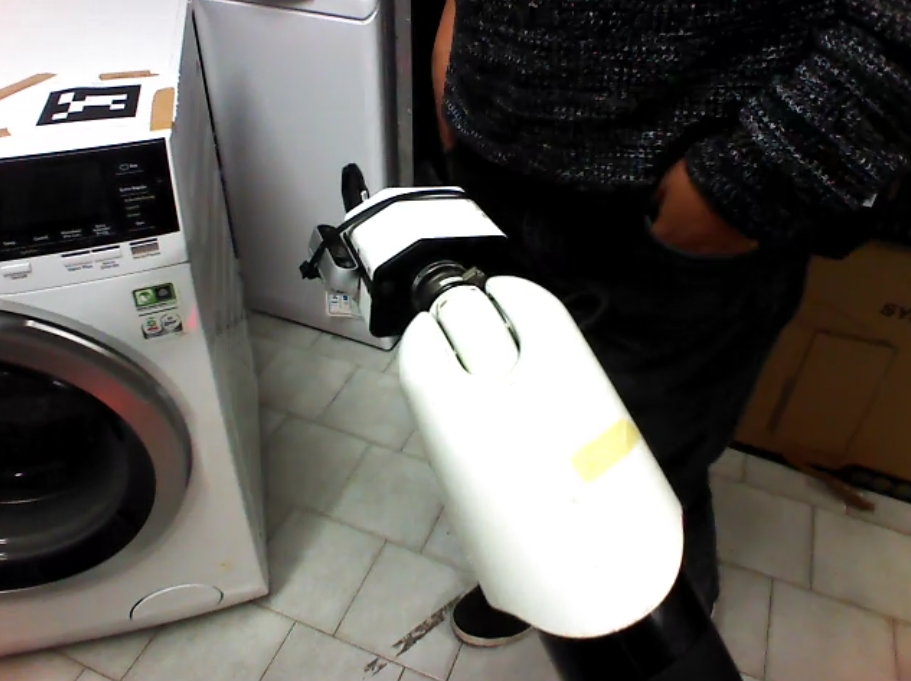}
 }
  \subfigure{
  \includegraphics[width=0.129\textwidth]{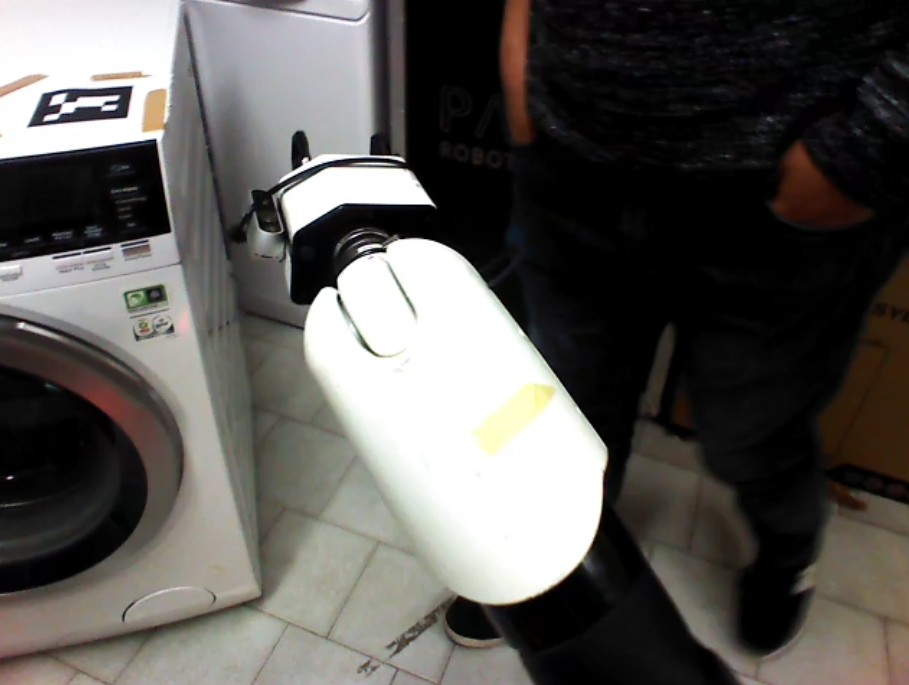}
 }
  \subfigure{
  \includegraphics[width=0.129\textwidth]{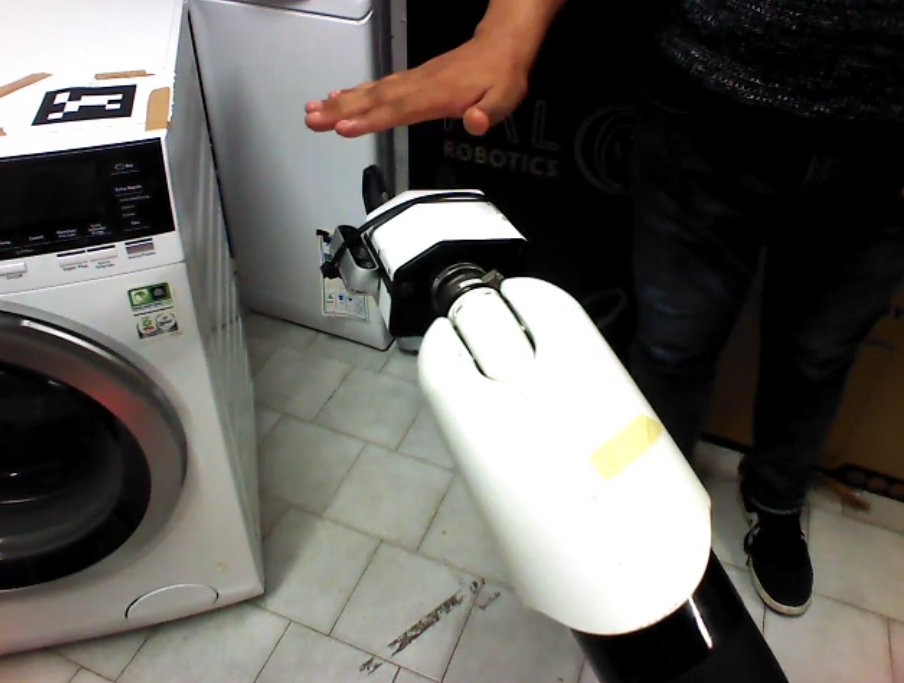}
 }
 \subfigure{
 \includegraphics[width=0.129\textwidth]{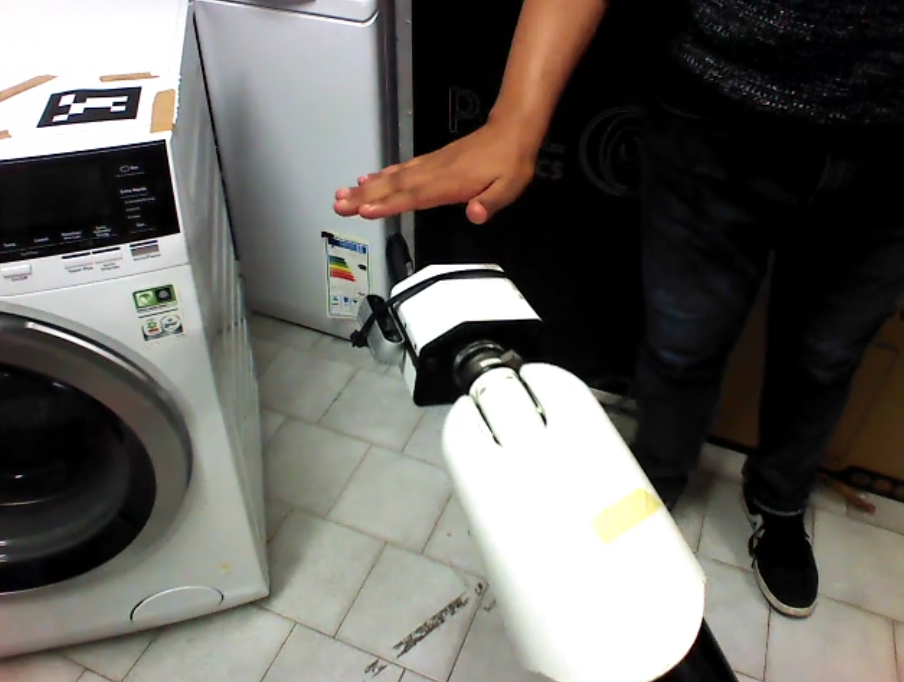}
 }
 \subfigure{
 \includegraphics[width=0.129\textwidth]{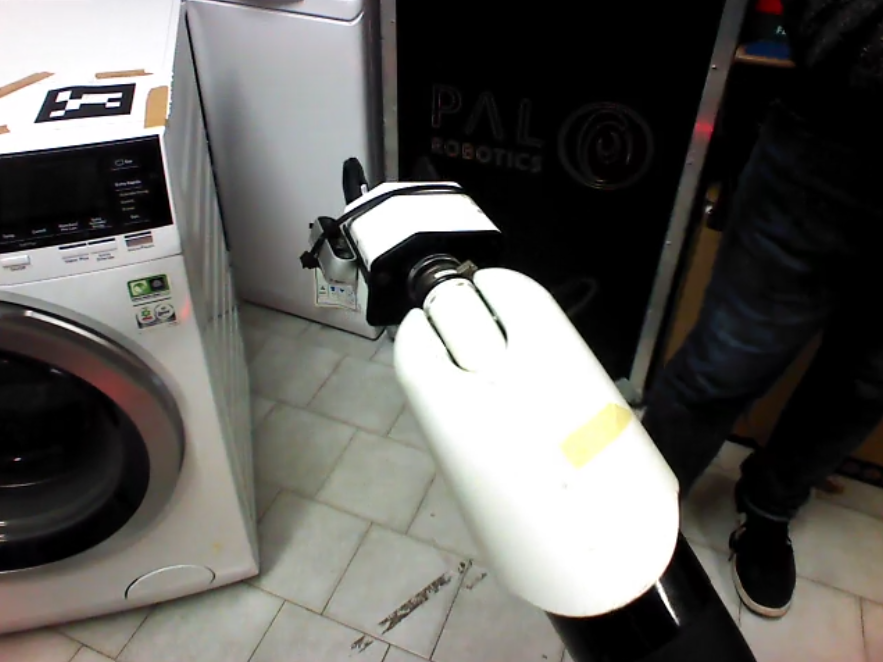}
 }
\subfigure{
\includegraphics[width=0.129\textwidth]{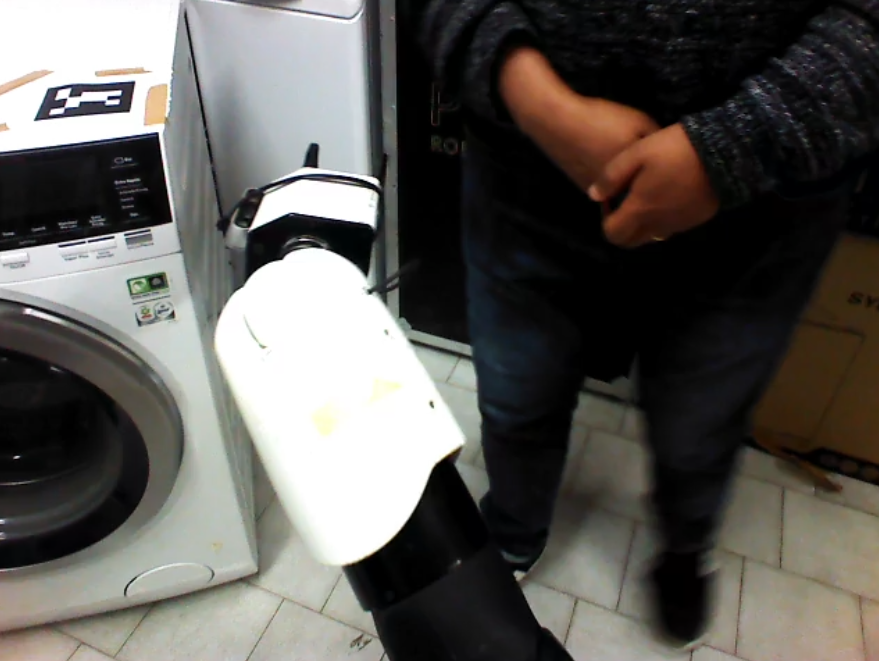}
 }
 \caption{ Experiment 1: The sequence of images top-down demonstrate real time collision avoidance from a fast moving person while tracking end-effector setpoint. Video demonstration of this Experiment (experiment-1) is included.}
  \label{fig:Tiago-person}
 \vspace*{-3mm}
\end{figure}

\begin{figure}[!ht]
 \centering
 \includegraphics[width=1\columnwidth]{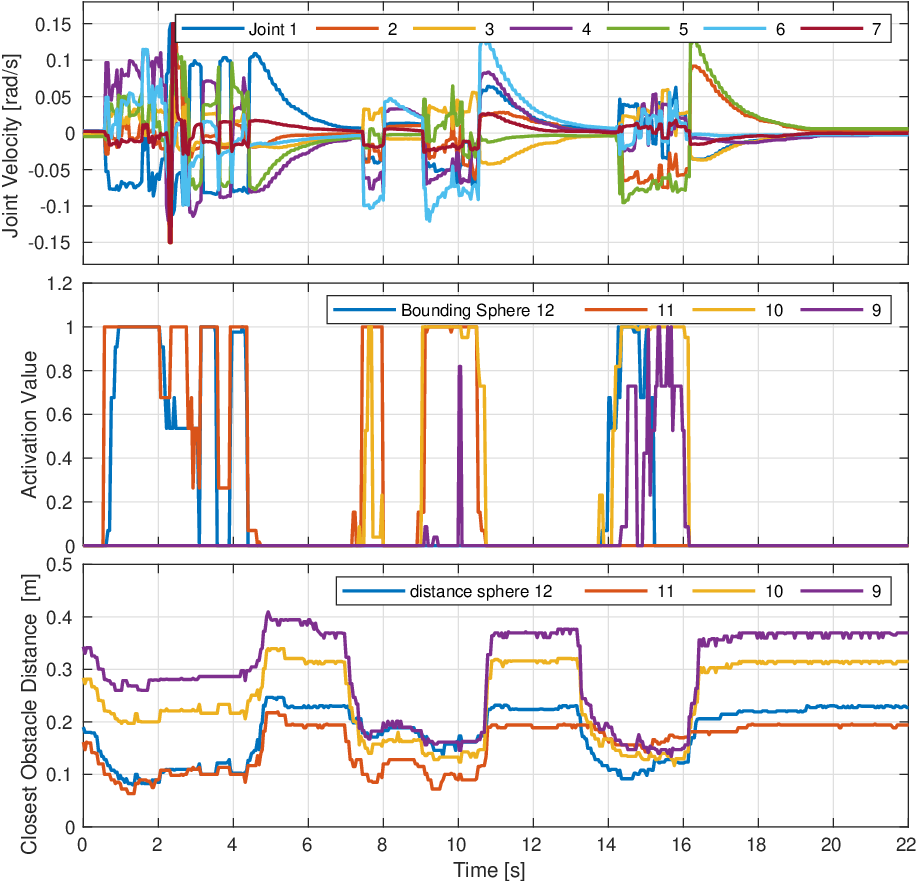} 
 \caption{Experiment 1: Commanded joint velocities, activation value and minimum obstacle distance during the collision avoidance from a moving person. 
 \label{fig:real-Tiago-wm-plot}}  
\end{figure}

The real time obstacle avoidance is demonstrated by selecting two test scenarios: the first test case is to interact with a dynamic scene formed by an approaching person while the robot is tracking a set point and second test scenario is to follow an end-effector trajectory while also avoiding dynamic obstacle. During our experiment, the PBA updates distance map at about 350$\,$Hz which is much higher than other robot-obstacles  distance evaluation  methods reported in literature. Even though scene point cloud is updated at 30 Hz, the control points  continue to move along with the robot link, therefore their corresponding minimum distance evaluation should be performed at higher rate. 
In the first scenario, the Tiago arm avoids collisions in real time from a dynamic obstacle, as shown in Fig.~\ref{fig:Tiago-person}. A person walking faster than an average human walk speed (approximately 1.4 m/s) approaches the robot from different directions. With reference to Fig.~\ref{fig:Tiago-person}, the first two images in the sequence show a person approaching the robot from the front, in the next three the person approaches the robot from the side, in the next two the person moves his hand close to the robot end effector from the top and finally full body approach is shown in the last two images. In all these cases, the robot successfully avoids collision. As the person retreats from the scene, the end-effector control task pushes the end-effector to the desired pose. The real time velocity command, obstacle avoidance task activation value and minimum distance associated to this maneuver is given in Fig.~\ref{fig:real-Tiago-wm-plot}. Note that the activation value and minimum obstacle distance for the \nth{12}, \nth{11}, \nth{10} and \nth{9} Bounding Spheres indicated in the plot of Fig~\ref{fig:real-Tiago-wm-plot} are associated to the last four spheres of the robot model and the remaining spheres are omitted because they remain deactivated throughout the experiment.  The collision avoidance activation matrix $A$ is also seen to get activated only when the person enters the scene and deactivated when the person exits, see Fig~\ref{fig:real-Tiago-wm-plot}.

\begin{figure}
    \centering
    \subfigure{
        \includegraphics[width=0.145\textwidth]{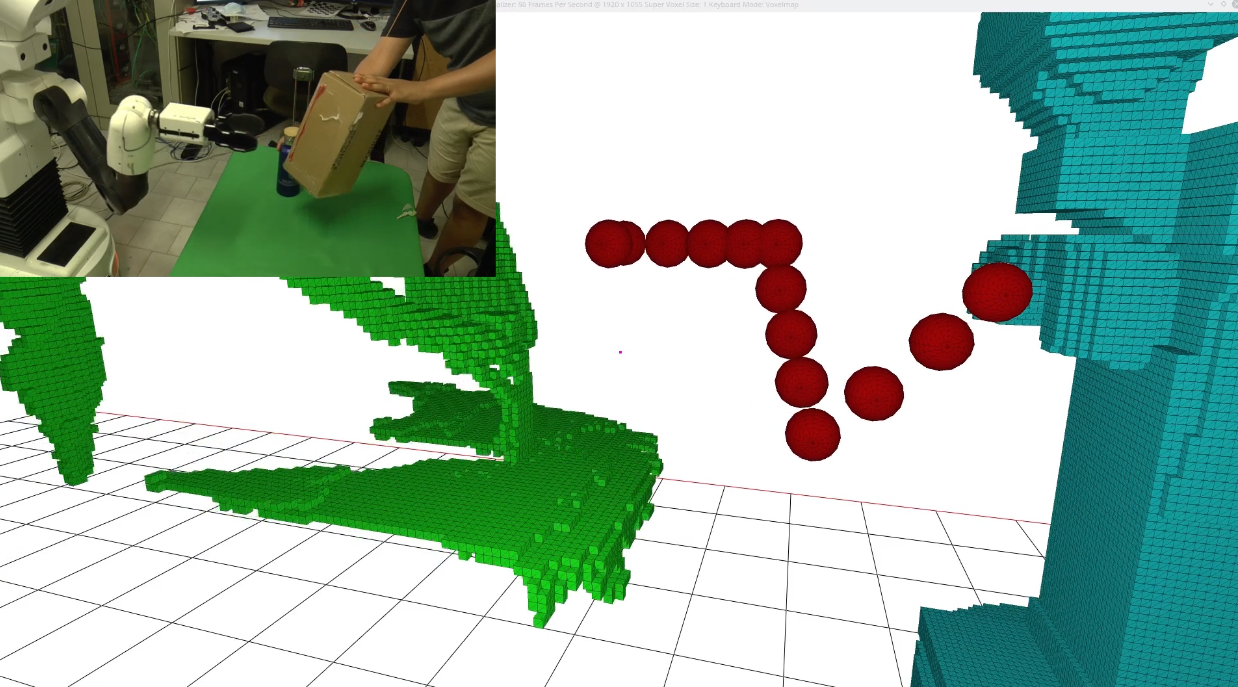}
    }
    \subfigure{
        \includegraphics[width=0.145\textwidth]{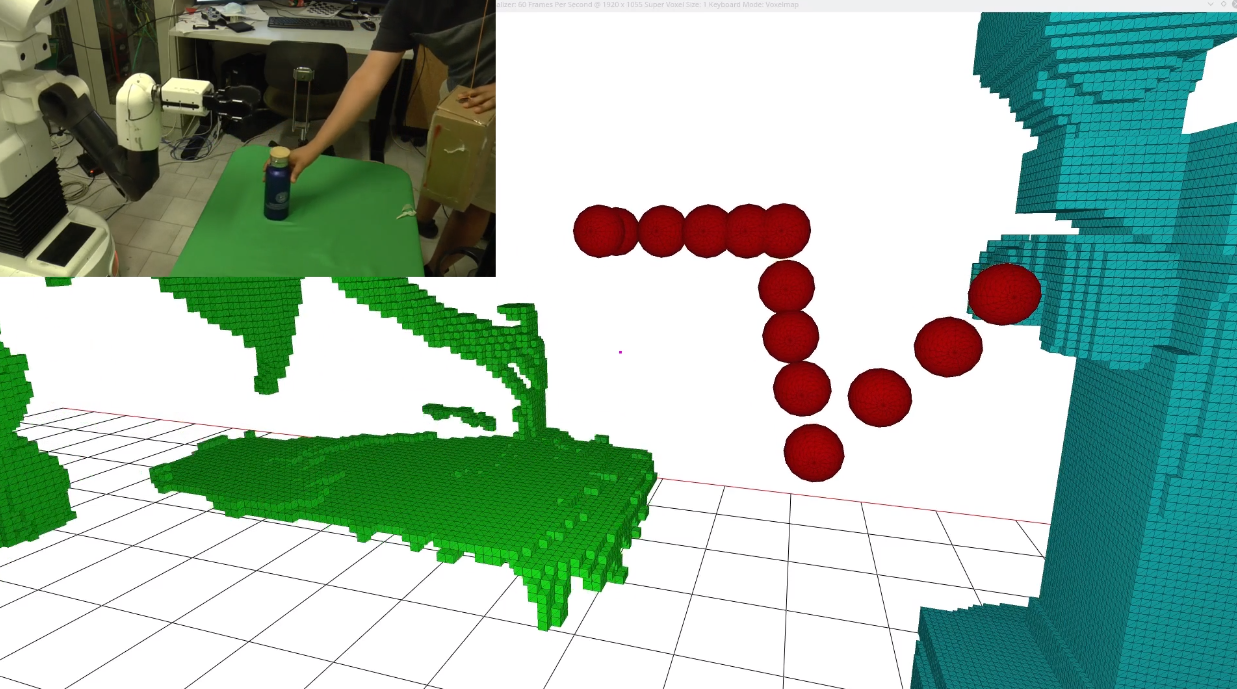}
    }
        \subfigure{
        \includegraphics[width=0.145\textwidth]{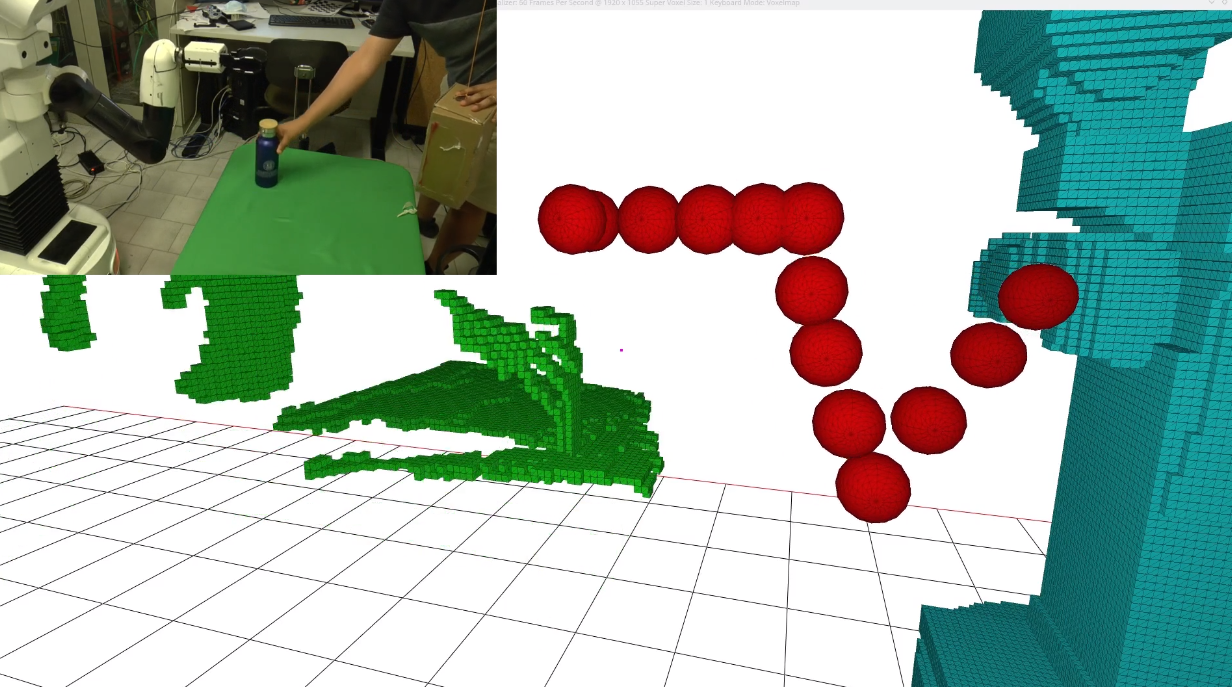}
    }

        \subfigure{
        \includegraphics[width=0.145\textwidth]{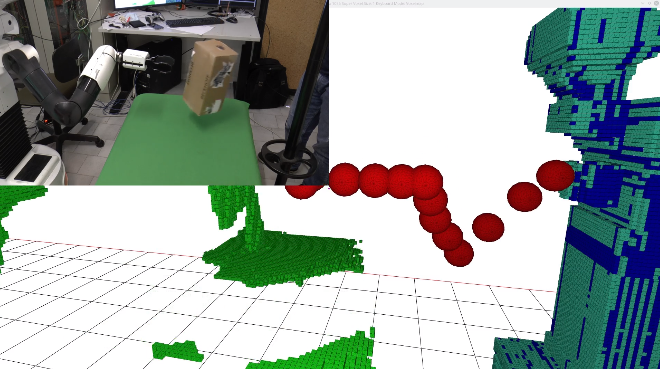}
    }
        \subfigure{
        \includegraphics[width=0.145\textwidth]{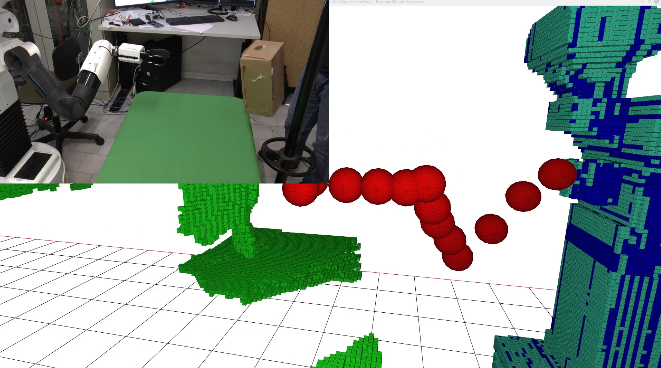}
    }
        \subfigure{
        \includegraphics[width=0.145\textwidth]{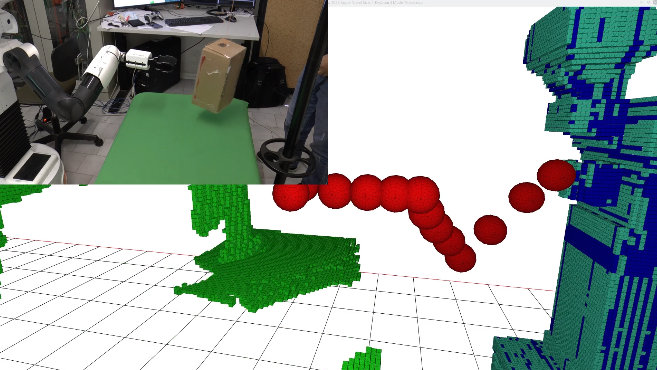}
    }
    \caption{Experiment 2: Image sequence for a motion through two way points with collision avoidance. An oscillating and spinning hanging box approaching the robot arm in the workspace. A video for experiment-2 is included with this work.}
    \label{fig:traj_obstacleavoidance}
    \vspace*{-3mm}
\end{figure}

In our second experiment, the robot end-effector follows a trajectory composed of two way-points, moving back and forth between them. A fast dynamic obstacle is formed by a box hanging through a thread, oscillating and spinning toward the robot, therefore requiring a fast reaction. As shown in Fig.~\ref{fig:traj_obstacleavoidance} and  the accompanying video\footnote{\url{https://drive.google.com/file/d/1qtlwCuN\_L7ACZUGftSaePC5BsfxW7OyP/view?usp=sharing}}, the robot avoids collision while also following end-effector trajectory as a lower priority task.



 \subsection{Task Oriented Regularization: Evaluation}
\begin{figure}
 \centering
 \includegraphics[width=20cm, height=6cm,
  keepaspectratio]{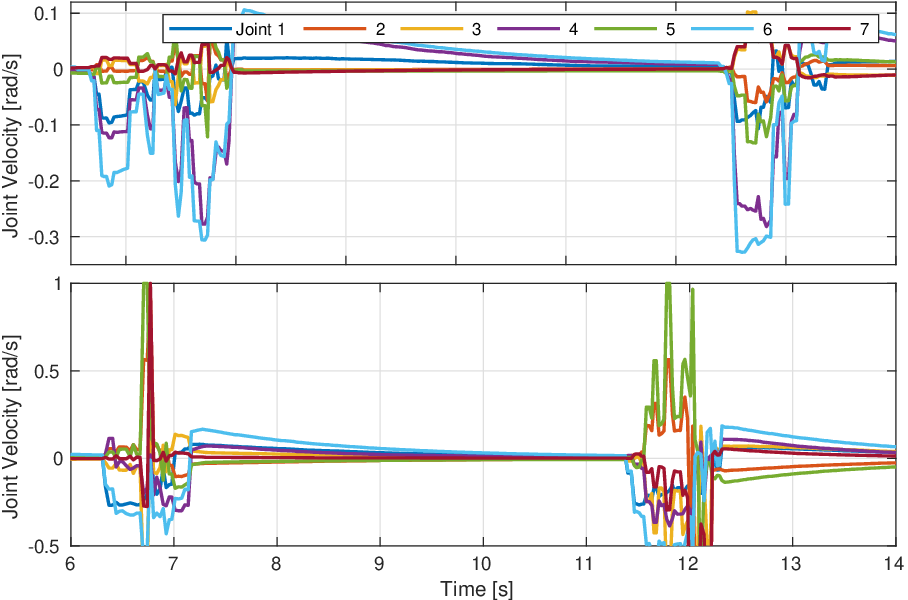}
 \caption{Experiment 3: Joint command: with Task oriented regularization (top) and without (bottom). A sharp velocity change is  observed when collision avoidance task is  activated and deactivated without task-oriented regularization.   
 \label{fig:regularization}}  
  \vspace*{-3mm}
\end{figure}

 To specifically evaluate concern with the practical discontinuity, an experiment (experiment 3) is performed in which the robot arm followed an identical trajectory with and without task oriented regularization.  Note that only collision avoidance tasks that are in transition are penalized, as proposed in the second term of eq.~\eqref{task_oriented}. The impact of the task-oriented regularization is highlighted when dynamic obstacles are involved in which the task is activated and deactivated rapidly. As shown in Fig.~\ref{fig:regularization}, the joint velocity command at the top contain regularization term which suppresses sudden velocity jumps during collision avoidance task transition.
\subsection{Comparison to related works}
The real time property is evaluated by examining the time elapsed to insert the scene point cloud to a GPU and compute distance map with various the map size.
As it can be seen from Table~\ref{pipeline time}, the rate of GPU based EDT depends linearly on the map size. But for a map sizes of practical interest (i.e. within the limit of the robot workspace) the distance map updates in the range of 200-500$\,$Hz, see Table~\ref{pipeline time}.
  A GPU based implementation comparable to our work is given in \cite{C144} which utilizes large set of vertices sampled from the mesh model of the robot as opposed to bounding spheres in our work to represent the robot. The reported time for distance field computation in \cite{C144} is at least 10$\,$ms (100$\,$Hz) for a map dimension of 2$\,$m$\,\times\,$2$\,$m$\,\times\,$1.8$\,$m, in addition to the 4 ms for potential field computation. Note that the approach given in \cite{C144} represents the environment in 2.5D and still our approach delivers higher performance. 
  
To evaluate the efficiency of the PBA compared to other GPU-compatible EDT algorithms, we considered two approaches from literature that could exploit modern GPU's. In the first case, we evaluate a naive approach in which the parallelism is limited to one thread per row. This is the baseline scenario achieved by setting the banding parameters to $m_1 = 1, m_2 = 1$ and $m_3 = 1$. Secondly, a Jumping Flood algorithm (JFA) in~\cite{rong2007variants} that works by creating outward ripple effect starting from the occupied voxels so that each voxel in the grid can decide which occupied voxel is it's closest one is also considered. Another algorithm comparable to JFA  is also presented in~\cite{schneider2009gpu} with time-work complexity of $\mathcal{O}(3^{d}N)$, where d is the size of each dimension in the grid. However, It is reported in~\cite{b99} that this algorithm performs slower in all scenarios considered compared to the PBA, therefore omitted from the comparison in Table~\ref{EDT comparisons}. In our comparison we considered three important factors: the level of parallelism (GPU utilization), algorithm run-time and work complexity and computation time for a practical map size. Although both PBA and JFA exploit GPU resource quite well, JFA has higher work-complexity that grows very fast with the map dimension as shown by the significant jump in time from $256^2\times128$ map size to $512^2\times128$ (Table~\ref{EDT comparisons}).

\begin{figure}
 \centering
 \includegraphics[width=0.9\columnwidth]{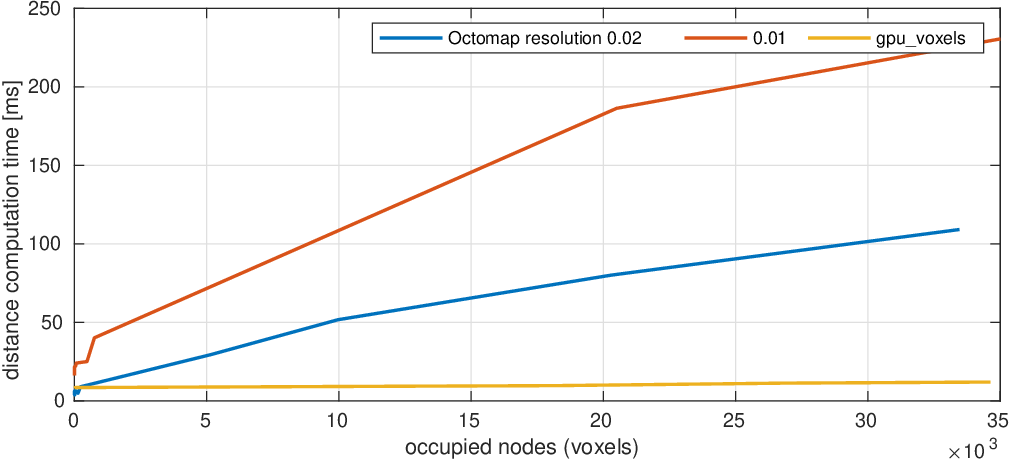}
 \caption{11 distance queries on Octomap as the number of occupied nodes increase shows significant change in the computation time (Red and Blue). On the other hand, time for distance queries are relatively unaffected as number of voxels occupied increases on GPU (Yellow).}
 \label{fig:occupancy}
\end{figure}

The comparison to algorithms whose distance computation relies on CPU based 3D representations such as Octomap and other methods utilizing approximations are given in Tab.~\ref{pipeline cpu}. Due to significant amount of latency in capturing and updating dynamic environment using a CPU based hierarchical data structures such as Octree (the underlying data structure of Octomap), which is usually less than 15$\,$Hz, comparison to our work is limited to static or very slowly varying scenes.
The work in \cite{C14} proposed two approaches to represent obstacles: the first utilizes Octomap directly while in the second approach they generate box collision objects from Octomap as an approximation. The results in Tab.~\ref{pipeline cpu} shows that, the approach in \cite{C14} is notably affected by the resolution of the Octomap as well as the robot model on which distance query is performed. Usage of high resolution Octomap in combination with mesh model for distance query leads to prohibitive computational time. On the other hand, for 11 bounding spheres distributed along the arm of Tiago robot, it takes about 20$\,$ms to complete distance computation from an Octomap of 2$\,$cm resolution. Although this time is acceptable in terms of real time execution, the performance deteriorates as the number of occupied nodes in the map rises, see Fig.~\ref{fig:occupancy}. Another approach proposed in \cite{C1} approximates the Octomap at different depth with sphere by replacing each occupied node with sphere. Here, it is important to note that the depth of query on the Octomap affects the computation time as well as the specific collision avoidance application. To elaborate this we performed a separate experiment (experiment 4) in which the robotic arm reaches inside a washing machine to perform inspection and grasping in the drum. To achieve this, An offline generated point cloud model of a complete washing machine is registered in to the scene using fiducial markers. For obstacle avoidance task during entering and existing the interior of an appliance, depth of 14 leafs on Octomap wasn't sufficient and going higher would incur larger computation time.
However, with the proposed method in this work, a one time EDT for this static scene was sufficient. As the robot arm moves minimum distance to the control points is extracted for collision avoidance.

\begin{table}[b]
\caption{work and time complexity comparison of GPU-based exact EDT computation algorithms for 3d voxel grid $N=n^3$}
\begin{tabular*}{\hsize}{@{\extracolsep{\fill}}llll@{}}
\toprule[1pt]\midrule[0.3pt]
\makecell{Factors} & \makecell{PBA\cite{b99}\\ $m_1$=$m_2$=1\\$m_3$=8} & \makecell{JFA\cite{rong2007variants}} & \makecell{PBA-naive\\ $m_1$=$m_2$=1\\$m_3$=1}\\
\toprule
\toprule
\makecell{GPU Utilization \\ (Threads and Bandwidth)} & Very high & Very high & under-utilized\\
time-work complexity & $\mathcal{O}(N)$ & $\mathcal{O}(N\log n)$ & $\mathcal{O}(N)$\\
\makecell{Run-time in (ms):\\ 256$\times$256$\times$128} & 3.18 & 26.67 & 3.52\\
\makecell{Run-time in (ms):\\ 512$\times$512$\times$128} & 7.782 & 114.47 & 11.271\\
\toprule 
\toprule
\label{EDT comparisons}
\end{tabular*}
\vspace*{-3mm}
\end{table}

\begin{figure}[t]
 \centering
  \subfigure{
  \includegraphics[width=0.13\textwidth]{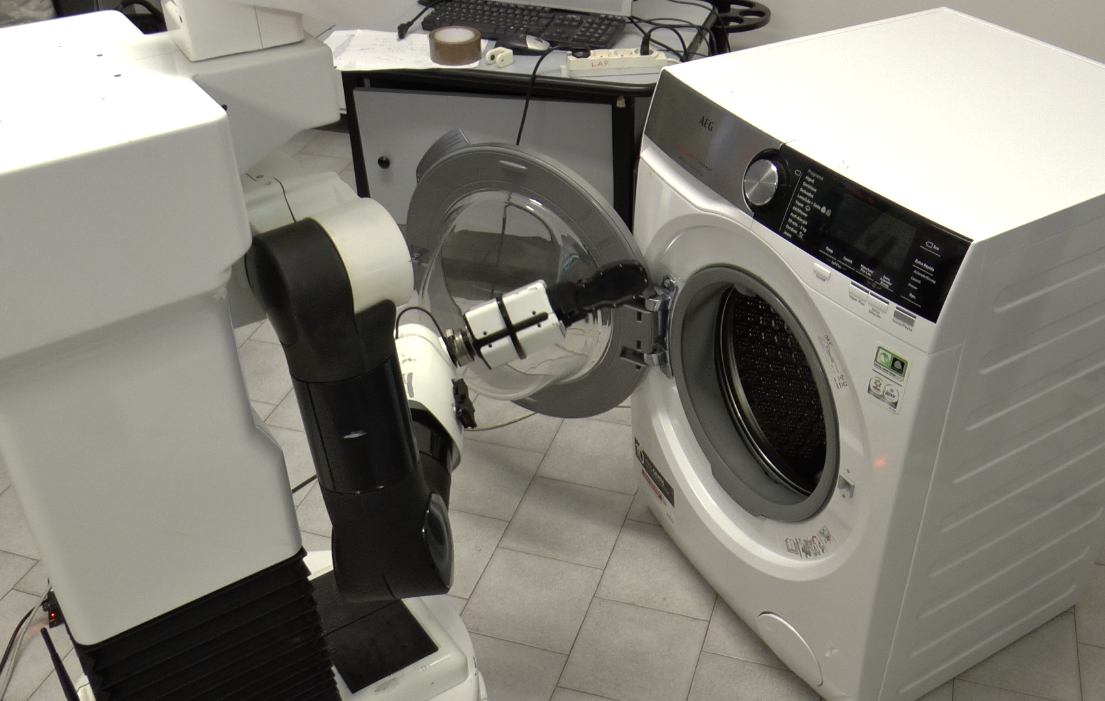}
 }
  \subfigure{
  \includegraphics[width=0.13\textwidth]{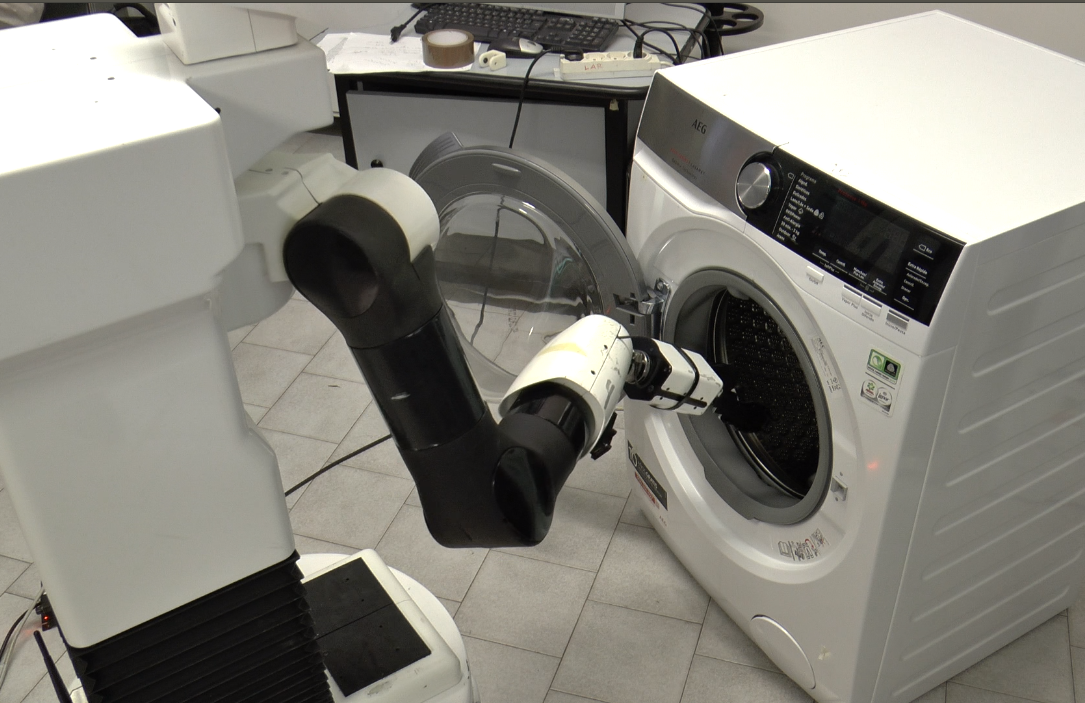}
 }
  \subfigure{
  \includegraphics[width=0.13\textwidth]{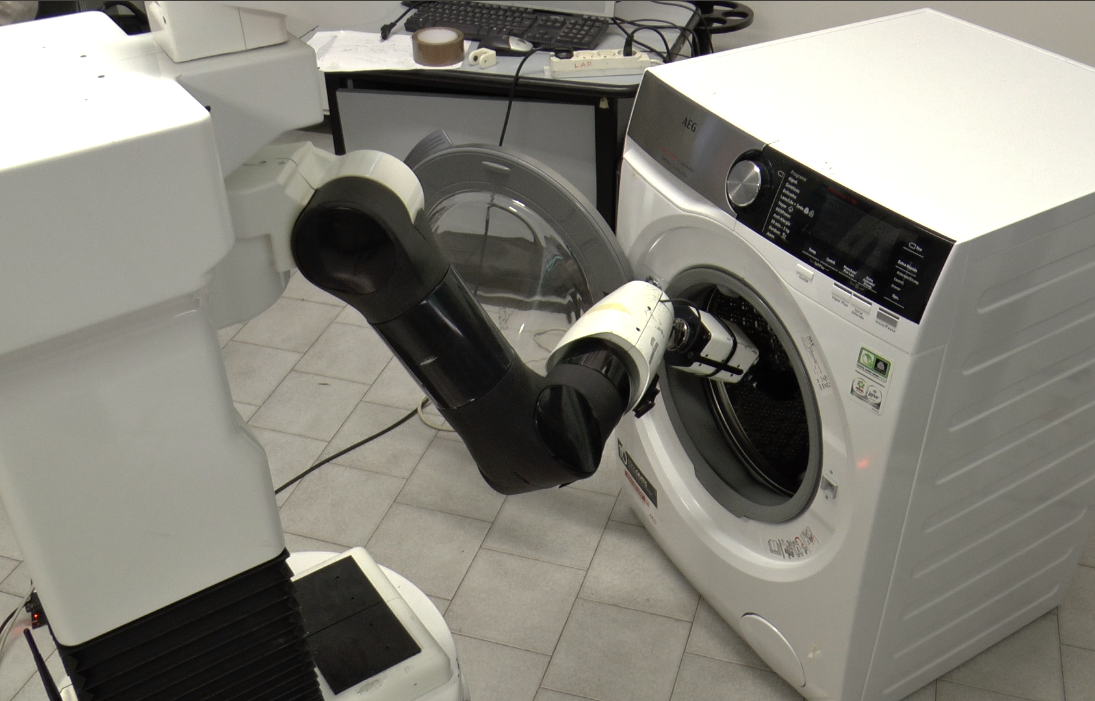}
 }
 \caption{ Experiment 4: Tiago robot arm entering the washing machine drum\label{fig:static_col_avoidance}}
 \vspace*{-3mm}
\end{figure}

\begin{table}
\caption{ Average minimum distance computation time (in ms) between robot and obstacle for CPU based models.}
\begin{tabular*}{\hsize}{@{\extracolsep{\fill}}llll@{}}
\toprule[1pt]\midrule[0.3pt]
& \multicolumn{3}{|c} {\textbf{Full Octomap obstacle representation \cite{C14}} } \\
\toprule 
\makecell{\textbf{Robot} \\
\textbf{model}} & \multicolumn{1}{|c} {1cm res.} & \multicolumn{1}{c} {2cm res.} & \multicolumn{1}{c} {4cm res.} \\
\toprule
 mesh Model & \multicolumn{1}{|c} {269.842} & \multicolumn{1}{c} {145.571} & \multicolumn{1}{c} {107.289} \\
  Capsules~\cite{safeea2019minimum} & \multicolumn{1}{|c} {162.446} & \multicolumn{1}{c} {113.105} & \multicolumn{1}{c} {78.059} \\
\toprule 
bounding
spheres & \multicolumn{1}{|c} {77.8657} & \multicolumn{1}{c} {19.7948} & \multicolumn{1}{c} {6.7741} \\
\toprule
\toprule 
& \multicolumn{3}{|c} {\textbf{Occupied Box approximation for obstacles \cite{C14}} } \\
\toprule 
& \multicolumn{1}{|c} {1cm res.} & \multicolumn{1}{c} {2cm res.} & \multicolumn{1}{c} {4cm res.} \\
\toprule
 mesh Model & \multicolumn{1}{|c} {142.446} & \multicolumn{1}{c} {48.1075} & \multicolumn{1}{c} {17.459} \\
\toprule 
bounding
spheres & \multicolumn{1}{|c} {131.719} & \multicolumn{1}{c} {36.56} & \multicolumn{1}{c} {8.3969} \\
\toprule
\toprule 
& \multicolumn{3}{|c} {\textbf{Occupied Spheres approximation for obstacles \cite{C1}} } \\
\toprule 
& \multicolumn{1}{|c} {14 depth} & \multicolumn{1}{c} {11 depth} & \multicolumn{1}{c} {9 depth} \\
\toprule
 mesh Model & \multicolumn{1}{|c} {142.446} & \multicolumn{1}{c} {48.1075} & \multicolumn{1}{c} {17.459} \\
\toprule 
bounding
spheres & \multicolumn{1}{|c} {17.724} & \multicolumn{1}{c} {11.661} & \multicolumn{1}{c} {9.351} \\
\toprule
\toprule 
\label{pipeline cpu}
\end{tabular*}
\vspace*{-3mm}
\end{table}

\subsection{Self-Collision Avoidance}
\begin{figure*}[t]
 \centering
 \subfigure{
  \includegraphics[trim=0 0 0 115,clip, width=0.12\textwidth]{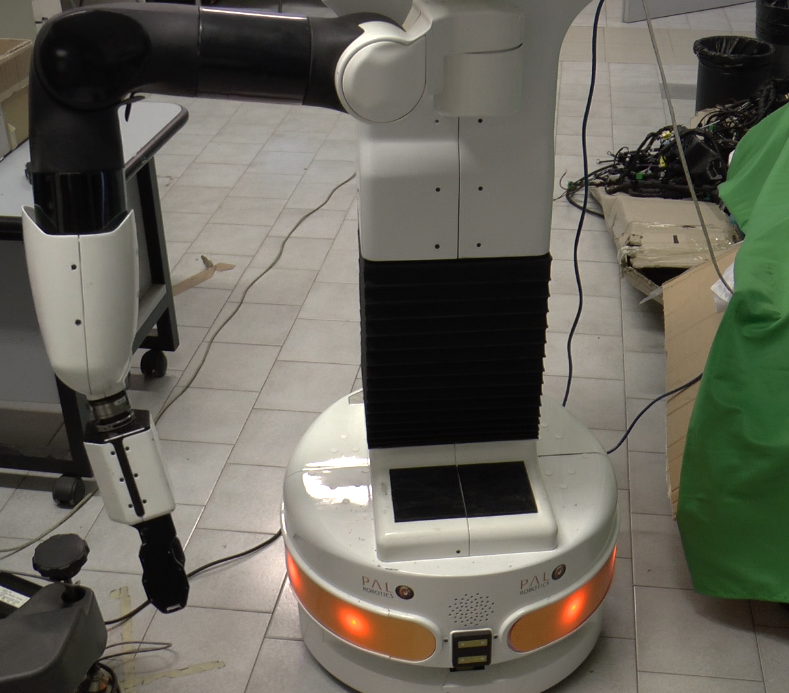}
 }
  \subfigure{
  \includegraphics[trim=0 0 0 135,clip, width=0.12\textwidth]{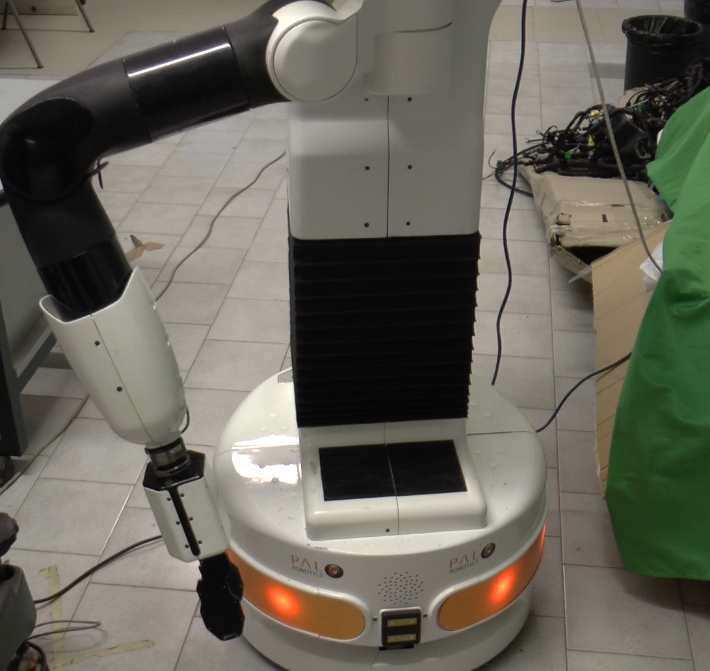}
 }
  \subfigure{
  \includegraphics[trim=0 0 0 15,clip, width=0.12\textwidth]{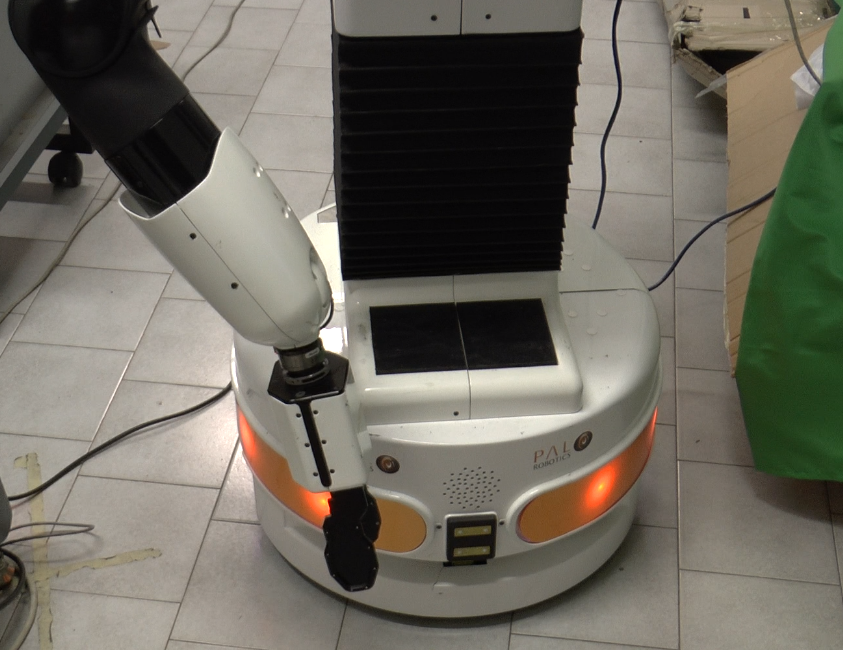}
 }
  \subfigure{
  \includegraphics[width=0.12\textwidth]{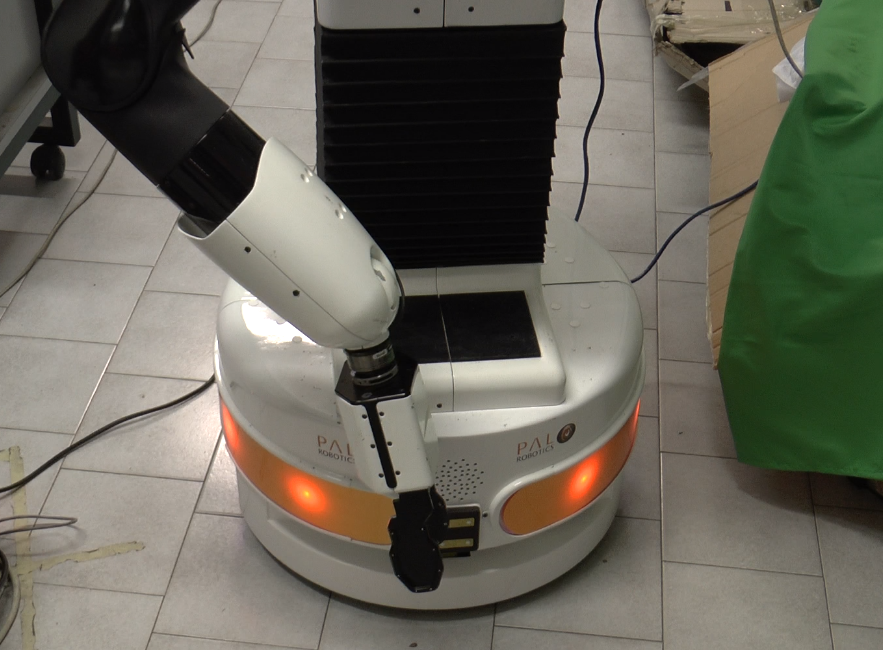}
 }
   \subfigure{
  \includegraphics[trim=0 70 0 0,clip, width=0.12\textwidth]{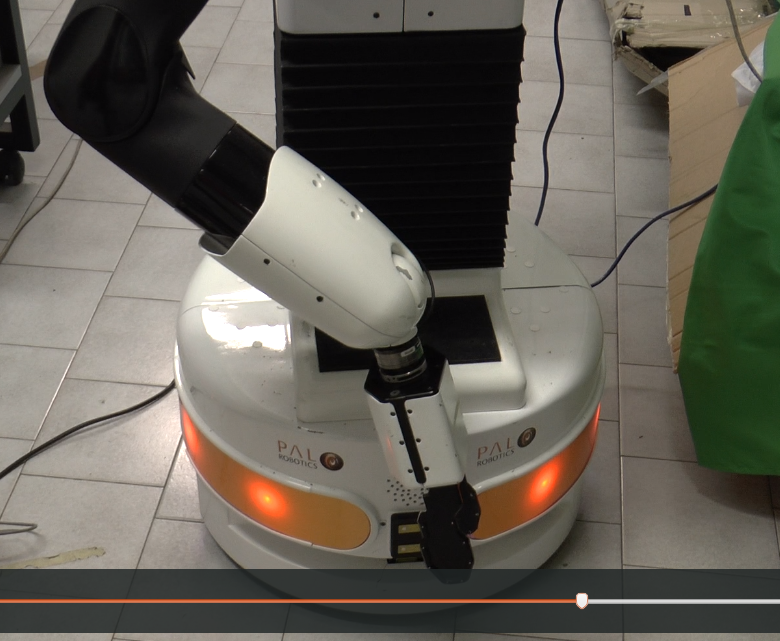}
 }
    \subfigure{
  \includegraphics[trim=0 40 0 0,clip, width=0.12\textwidth]{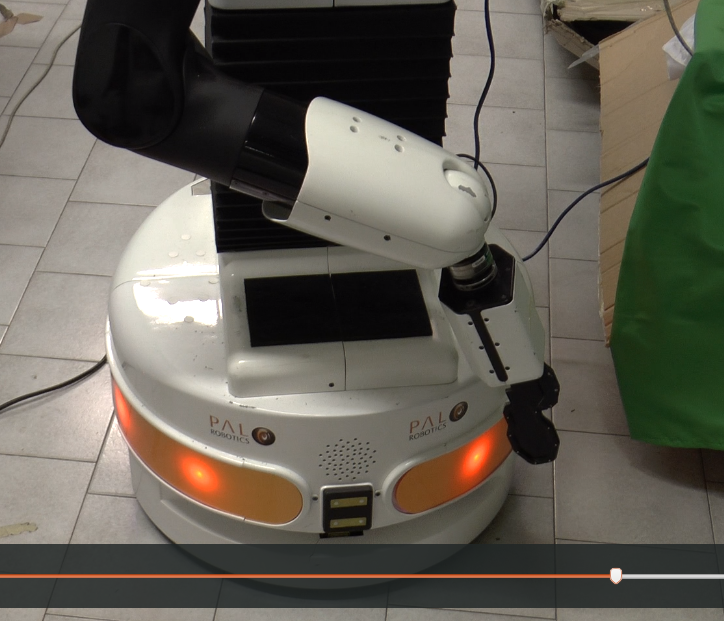}
 }
    \subfigure{
  \includegraphics[trim=0 40 0 0,clip,width=0.12\textwidth]{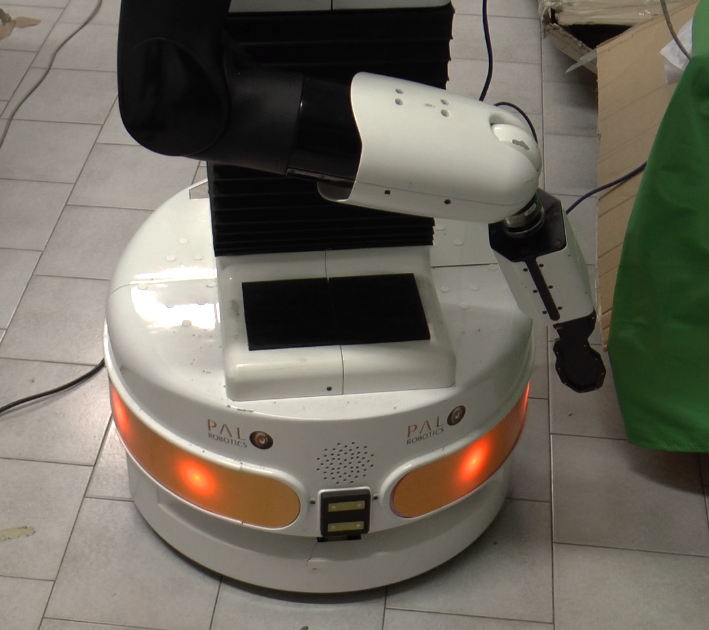}
 }
  \\
  \subfigure{
  \includegraphics[trim=0 0 0 45,clip, width=0.13\textwidth]{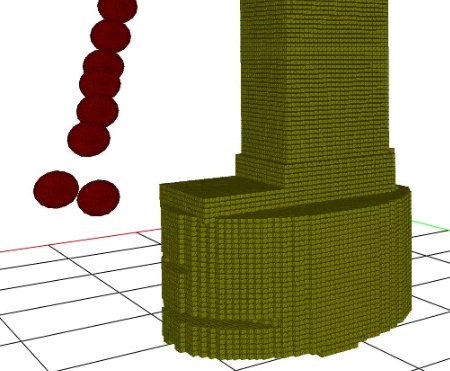}
 }
  \subfigure{
  \includegraphics[trim=0 0 0 35,clip, width=0.13\textwidth]{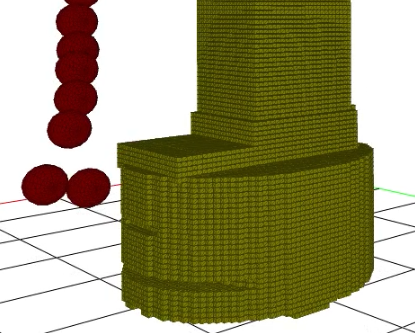}
 }
  \subfigure{
  \includegraphics[trim=0 0 0 0,clip, width=0.13\textwidth]{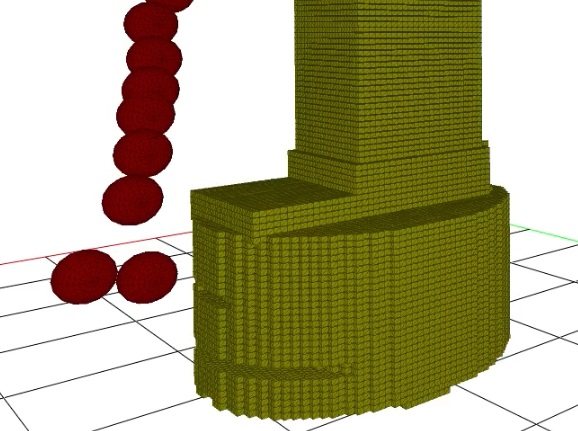}
 }
  \subfigure{
  \includegraphics[width=0.13\textwidth]{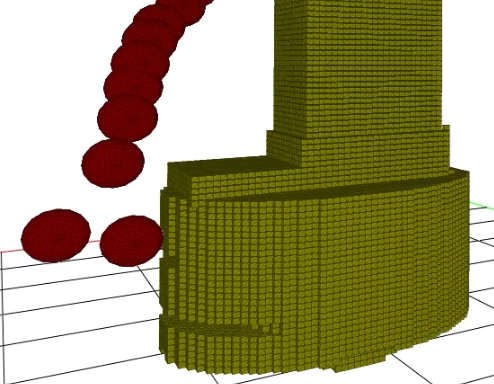}
 }
   \subfigure{
  \includegraphics[trim=0 0 0 0,clip, width=0.09\textwidth]{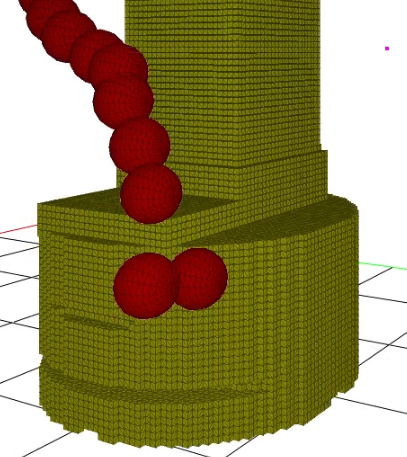}
 }
   \subfigure{
  \includegraphics[trim=0 0 0 0,clip, width=0.09\textwidth]{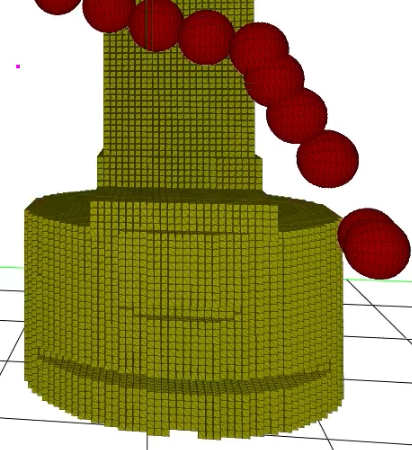}
 }
    \subfigure{
  \includegraphics[trim=0 0 0 0,clip, width=0.09\textwidth]{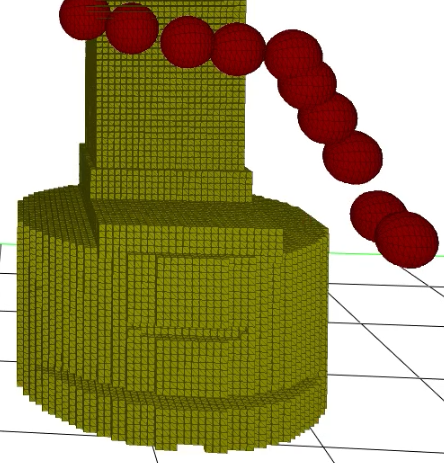}
 }
  \caption{Experiment 5: Sequence of images (top) showing The arm navigating around the rim of the circular base platform to Avoid self collision while also moving towards a set point. Visualization of  robot body as a collision object (bottom sequence).\label{fig:self_col_avoidance}}
 \vspace*{-3mm}
\end{figure*}

This  experiment demonstrates Tiago arm self collision avoidance with the entire body in real time. The distance between the bounding spheres of the arm and the rest of the robot part is computed and a self-collision avoidance velocity is generated according to Alg.~\ref{algo:self} and eqs.~\eqref{eq:selftaskvar}-\eqref{eq:selftaskjac} as Tiago arm navigate from initial pose to a final pose along the rim of the circular base, as shown in the image sequence shown in Fig.~\ref{fig:self_col_avoidance}.

\begin{figure}
 \centering
 \includegraphics[width=1\columnwidth]{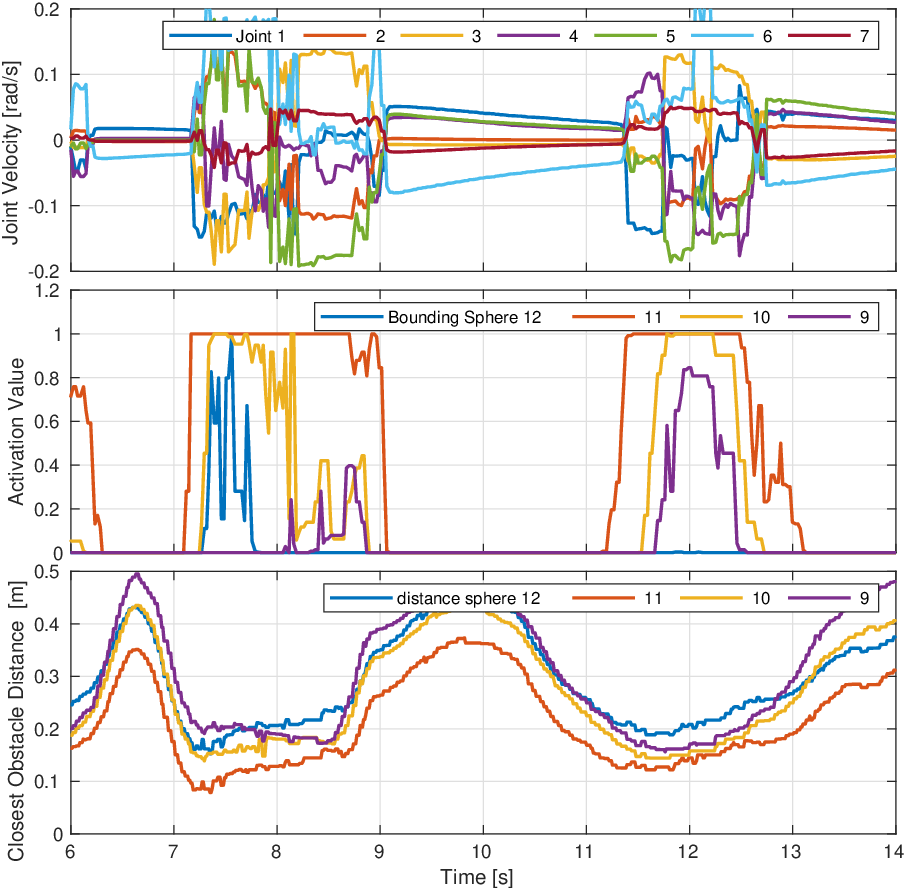}
 \caption{Self collision avoidance: joint velocity, activation and minimum distance during Experiment 5 shown in Fig~\ref{fig:self_col_avoidance}. 
 \label{fig:self_col_avoidance_plot}
}  
\end{figure}

The minimum distance of the bounding spheres on the gripper and the corresponding activation matrix and joint velocities shown in Fig.~\ref{fig:self_col_avoidance_plot} demonstrates that the robot arm maintained the desired threshold distance during the movement toward the set-point.

\section{Conclusions} \label{conclusion}
We have presented a unified real time self and obstacle  collision avoidance method that is based on kinematic task priority control. The main idea in the proposed algorithm is to perform minimum distance computation for both external obstacle and self collision on a GPU at a very high rate so that the robot is able to react instantaneously to avoid collision from dynamic obstacles as well as it's own link. Multiple experiments on our mobile robot platform Tiago demonstrated the real time effectiveness  of the method. A comparison to related works on reactive collision avoidance has also been given showing that the level of occupancy, resolution and depth of query of the environment representation significantly affect their performance.

Future work will involve estimation of the velocity of the dynamic obstacle which will improve the obstacle avoidance task. In addition we also intend to introduce safety constraints for human interacting with robot. 

\bibliographystyle{IEEEtran}
\bibliography{refs}


\end{document}